\documentclass[]{article}

\usepackage{arxiv}

\usepackage[utf8]{inputenc} 
\usepackage[T1]{fontenc}    
\usepackage{hyperref}       
\usepackage{url}            
\usepackage{booktabs}       
\usepackage{amsfonts}       
\usepackage{nicefrac}       
\usepackage{microtype}      
\usepackage{lipsum}		
\usepackage{graphicx}
\usepackage{natbib}
\usepackage{doi}
\usepackage{algorithm}
\usepackage{algpseudocode}
\usepackage{caption}

\usepackage{url,hyperref,lineno,microtype,subcaption}
\usepackage[onehalfspacing]{setspace}
\usepackage{caption}
\usepackage{algpseudocode}
\usepackage{algorithm}
\usepackage{float}
\usepackage{booktabs}

\usepackage{url,hyperref,lineno,microtype,subcaption}
\usepackage[onehalfspacing]{setspace}
\usepackage{caption}
\usepackage{algpseudocode}
\usepackage{algorithm}
\usepackage{float}
\usepackage{booktabs}
\usepackage{tabularray}
\usepackage{siunitx} 
\usepackage{arydshln}

\usepackage{nicematrix}
\DeclareCaptionType{equ}[][]

\usepackage{authblk}
\author[1,2,†]{Noah Barrett}
\author[1,2,†,*]{Zahra Sadeghi}
\author[1,2]{Stan Matwin}
\affil[1]{Facutly of Computer Science, Dalhousie University, Halifax, Canada}
\affil[2]{Institute for Big Data Analytics, Halifax, Canada}
\affil[*]{Correspondence: zahras@dal.ca}

\begin{document}
\onecolumn

\title {Evolutionary Augmentation Policy Optimization for Self-Supervised Learning} 

\date{\vspace{-5ex}}

\maketitle
\def\thefootnote{$\dagger$}\footnotetext{These authors contributed equally to this work}

\begin{abstract}
Self-supervised Learning (SSL) is a machine learning algorithm for pretraining Deep Neural Networks (DNNs) without requiring manually labeled data. The central idea of this learning technique is based on an auxiliary stage aka pretext task in which labeled data are created automatically through data augmentation and exploited for pretraining the DNN. However, the effect of each pretext task is not well studied or compared in the literature. In this paper, we study the contribution of augmentation operators on the performance of self supervised learning algorithms in a constrained settings. We propose an evolutionary search method for optimization of data augmentation pipeline in pretext tasks and measure the impact of augmentation operators in several SOTA SSL algorithms. By encoding different combination of augmentation operators in chromosomes we seek the optimal augmentation policies through an evolutionary optimization mechanism. We further introduce methods for analyzing and explaining the performance of optimized SSL algorithms. Our results indicate that our proposed method can find solutions that outperform the accuracy of classification of SSL algorithms which confirms the influence of augmentation policy choice on the overall performance of SSL algorithms. We also compare optimal SSL solutions found by our evolutionary search mechanism and show the effect of batch size in the pretext task on two visual datasets. 

\tiny
\keywords{ Self-supervised Learning (SSL), Data Augmentation, Evolutionary Algorithms, Deep Learning, Deep Neural Networks, Optimization, Explainability, Artificial Intelligence} 
\end{abstract}

\section{Introduction}
The supreme power of Machine Learning algorithms is founded on supervised learning techniques. However, while these algorithms have shown to be tremendously successful in solving classification problems, they remain entirely dependent on a large corpus of manually annotated data. As a result, the whole process of learning is not autonomous and tends to be influenced by the error of annotation. 
Self-supervised learning (SSL) has been introduced in response to tackling these limits by making the possibility of training DNNs without relying on labelled data. 
The core of SSL is based on an auxiliary phase, aka pretext task which focuses on pretraining networks using automatically generated labeled data. One crucial aspect of pretext tasks is transforming unlabeled images using augmentation operators in order to produce labeled samples. 
Although many different SSL paradigms have been developed, there is no specific investigation about their level of effectiveness and usefulness. Moreover, there is no specific study to investigate the effect of augmentation operators on the performance of each of the proposed self-supervised algorithms. 
 In this study, we show that the choice of augmentation operators can bear an impact on the achievable performance of the downstream problem. Through a series of experiments, we seek to measure the influence of different augmentation policies.
 We formulate an evolutionary search algorithm in order to optimize the transformation settings that lead to best outputs. To this end, we focus on four state-of-the-art SSL algorithms that have indicated high performance on various datasets. 
This paper encompasses the following original contributions. We first investigate the best combination of augmentation operators in four state-of-the-art SSL algorithms through an evolutionary learning algorithm. Then, we compare the performance of these SSL methods before and after the optimization process. We find and evaluate the best augmentation operators for each SSL method across two different datasets. In the end, we run explanatory analysis over the optimized chromosomes to understand the impact and importance of augmentation operators for each SSL algorithm and dataset.
This paper is organized as follows: We provide an overview about self-supervised learning, auto augmentation and search optimization algorithms in section 2. In section 3, implementation details are mentioned. Then, in section 4, our method is described. The experiments and results are provided in section 5. The explainability methods are presented in section 6. Finally, section 7 concludes the paper.
 
\section{Related work}
In this section, we provide a literature survey about related prior work in three subsections. First, we introduce the paradigm of self-supervised learning algorithms and review ongoing research in this field. Secondly, we go over recent advances in Auto Augmentation. Finally, we provide an overview about intelligent search algorithms.

\subsection{Self-Supervised Learning}
Self-supervised learning has increased attention in machine learning due to its ability to handle unlabeled data more efficiently than the traditional unsupervised learning. This is achieved by designing a new phase of learning known as pretext task, which is an attempt for solving an auxiliary classification task. The auxiliary task is focused on classifying the data into self-labeled categories in order to obtain a pretrained model with rich feature representation about the underlying structure of data. 
Various approaches have been proposed for designing the pretext tasks, which can be classified into patch-based, instance-based, and video-based methods. Patch-based methods aim to learn the relative position between patches of an image \cite{doersch2015unsupervised}, while instance-based methods utilize the whole image and take advantage of an augmentation task such as an affine transformation \cite{doersch2015unsupervised} or colorization \cite{isola2016colorful}. In contrast, video-based methods design a task that can take advantage of the sequential ordering of frames in a video clip \cite{misra2016shuffle}. 
Self-supervised learning can also be categorized into generative or contrastive models. Generative modeling approaches aim to reconstruct the input through the deployment of autoencoders \cite{pathak2016context}, and GANs \cite{donahue2016adversarial}, while the contrastive approach includes the attempts to increase the discrimination between different images and decrease it for similar ones by applying contrastive learning \cite{oord2018representation}. Self-supervised learning is highly associated to contrastive models, while the generative approach is often considered as an unsupervised learning scheme. 
A variety of different approaches have been proposed in order to solve the pretext task, which is highly focussed on different forms of augmentation, such as spatial location and relative positions of patches \cite{noroozi2016unsupervised}\cite{dosovitskiy2014discriminative}, image transformation including rotation \cite{gidaris2018unsupervised}, altering global statistics while preserving local statistics \cite{jenni2020steering}, data and color jittering, painting and coloring \cite{isola2016colorful}. 
Given the fact that such pretext tasks are intrinsically different, it is very intriguing to investigate the effect of pretraining process of pretext task in quest of discovering performance differences in downstream tasks. 
In this paper, we focus on comparing four state-of-the-art SSL algorithms, namely, BYOL \cite{grill_bootstrap_2020}, simsiam \cite{chen2021exploring}, NNCLR \cite{dwibedi2021little} and SWAV \cite{caron_unsupervised_2021}. All of these SSL algorithms share fundamental similarities in their architecture and underlying methodology. Specifically, all algorithms employ a Siamese architecture at its core. Siamese architectures consist of a shared input network which feeds into two separate neural networks that share weights, and an architecture that is analogous to siamese twins in nature. Conventionally, Siamese networks are used for images where the inputs are different images, and the actual comparability of images are determined in a supervised manner \cite{chen2021exploring}. BYOL, SimSiam, NNCLR, and SwAV share the general notion of contrastive learning using a siamese architecture, however the specific details of implementation and theoretical motivation creates significant differences in the four algorithms. It is of great importance to note that the authors of these four algorithms highlight, to varying degrees, their invariability to augmentation. The proposed work looks into the margins of these findings and expose that while there is an astounding invariance to augmentation strategies in the proposed algorithms overall, there still is room for improvement for understanding the effect of augmentation in contrastive learning. 
The key idea of BYOL is that from a given representation, called the target representation, it is possible to train an enhanced representation,i.e., the online representation. BYOL works iteratively, to improve the learned representation by using a slow moving average of the online network as the target network \cite{grill_bootstrap_2020}. BYOL does not require negative samples, and claims to be insensitive to batch sizes and the types of augmentations used. BYOL's ability of not requiring a large batch sizes, negative pairs, as well as having a novel level of robustness to augmentations, was what made it a breakthrough algorithm at the time of its release. BYOL's robustness to augmentation is in comparison to one of the earliest forms of contrastive SSL SimCLR \cite{chen_simple_2020}; it is shown that BYOL suffers a much smaller performance drop than SimCLR when only using random crops. 
SimSiam is an SSL method that opts for simplicity and highlights the significance of the Siamese network architecture of SSL methods. SimSiam shows that the shared weight configuration is key to many core SSL algorithms, and a simple stop gradient approach can be used to prevent collapse \cite{chen2021exploring} which is a common issue  in many algorithms employing Siamese networks. Collapse occurs when the model maps every image to the same point, maximizing similarity when comparing, but in turn not learning any important information about problem at hand. SimSiam provided surprising evidence that meaningful representations can be learned without the use of negative sample pairs, large batches and momentum encoders. The first two of these were already unnecessary in the earlier work shown in BYOL, however, the momentum encoder component was thought of as a crucial component to its avoidance of collapsed solutions. Perhaps, the most remarkable discovery produced by SimSiam could be that a simple stop gradient is all that is needed to successfully train a Siamese architecture. 
Compared to BYOL and SimSiam, SwAV provides a novel mechanism for learning. Following the vanilla approach of contrastive learning, BYOL and SimSiam learn by predicting the "closeness" of two views of a sample, this is a core concept to Siamese networks and their impactful discoveries in representation learning. Simply put, SwAV learns by computing a code from an augmented version of an image and then predicts this code from other augmented versions of the same image. The method simultaneously clusters data while enforcing consistent cluster assignments between different views \cite{caron_unsupervised_2021}. Additionally, SwAV introduces a novel augmentation multi-crop, which allows for more memory efficient comparisons by utilizing a small set of large crop windows and a larger set of small crop windows. This means, despite employing a Siamese architecture, SwAV is not actually a form of contrastive learning. However, SwAV, at its core, still aims to learn a strong latent representation of the problem at hand. The use of this alternate approach yields a new breakthrough unfound by BYOL and SimSiam, which is the ability to significantly reduce the number of pairwise feature comparisons, reducing the computational and memory requirements of the algorithm. 
Compared to the other discussed SSL methods, NNCLR introduces a very novel concept with respect to the selection of positive pairs. Rather than deriving positive samples for an image via augmentation, they show that it is possible to use the nearest neighbours of an image of interest in the given data space as positive samples. Like BYOL and SimSiam, NNCLR is a contrastive learning method, which, similar to SimSiam, employs a very simple architecture. The novelty is in the selection of positive samples from the given dataset rather than derivation from augmentation. NNCLR samples positive samples by computing the nearest neighbours of a given sample in the learned latent space of the dataset. This approach provides more semantic class-wise variations rather than predefined transformations which tend to provide more geometric information \cite{dwibedi_little_2021}. Alike all the above-discussed approaches, NNCLR finds that it is not very reliant on data augmentation. 

All four discussed algorithms share the use of a Siamese architecture; SimSiam can be thought of as the simplest version of these learning approaches using a Siamese architecture. BYOL adds a slight sort of complexity to the learning method by employing a slow-moving average approach for the target component of the Siamese network. Like SimSiam, NNCLR also opts for simplicity in its architecture, however, it refrains from employing only augmentation to generate positive samples, and instead, determines the nearest neighbours of a given image to select positive samples. SwAV differs from all three methods in that it utilizes a swapped prediction mechanism for learning rather than the typical contrastive mechanism. All four algorithms have recently shown remarkable breakthroughs in the self-supervised computer vision world, and for this reason they have been chosen as focal points in our work. The four algorithms all claim to have a varying degree of invariance to data augmentation based on empirical findings. However, the contrastive mechanism of BYOL, SwAV and SimSiam, rely solely on  data augmentation. Given the limited discussion on the effect of data augmentation, the question of its impact to the different SSL algorithms still remains. In this work, we hypothesize that by picking the best set of augmentation operators, we can boost the performance of SSL algorithms.

\subsection{Auto Augmentation}
Data Augmentation is a regularization technique that has been mainly developed for dealing with overfitting issue in training DNNs by increasing the diversity of training examples \cite{shorten2019survey}. Recently, this approach has been adopted in contrastive self-supervised learning procedure for creating additional positive samples while preserving the semantic content of observations \cite{von2021self}. As mentioned in previous section, the positive samples will be discriminated against negative samples in a training procedure to obtain a rich embedding that can be used for learning different tasks. While data augmentation is the core basis of self-supervised learning, little attention has been placed over understanding its impact and there are not many investigations about the theoretical success of augmentation tasks. To the best of our knowledge there is no systematic study to assess the effect of employing different augmentation policies in self-supervised learning. Some authors have pointed out the important role of transformation operators in self-supervised feature learning performance. Jenni et al., argued that transformation function selection is data-specific and hence each dataset might benefit from a different set of transformations \cite{jenni2020steering}. In addition, different tasks could require tailored augmentation techniques. For example, for anomaly detection a novel proxy task has been proposed \cite{li2021cutpaste}. 
Along the recent endeavors in AutoML which aim to automate the Machine Leaning procedure, auto augmentation methods are developed in order to seek the optimal augmentation strategies for training deep neural networks to improve their performance. AutoAugment can be regarded as one of the pioneering approaches for designing an automatic data augmentation learning procedure \cite{cubuk2019autoaugment}. This approach is based on employing reinforcement learning to explore search space of augmentation policies. In \cite{lim2019fast} the auto augmentation procedure is accelerated by matching the density between training and validation data. PBA \cite{ho2019population} is another approach for that uses a hyperparameter search algorithm to learn a schedule of augmentation polices. 

\subsection{Search optimization algorithms}
Search optimization algorithms are one of the core components for the tremendous success of machine learning techniques and they are increasingly exploited in fundamental problems of parameter and hyperparameter tuning and model selection \cite{brazdil2022metalearning}. An optimization algorithm seeks to adjust the configuration of machine learning algorithms by minimizing a cost function such that it will lead to accurate estimation and prediction. Through iterative search algorithms, the best configuration that best describes a distribution of observable data can be obtained. There are a variety of search optimization techniques which are employed for machine learning tasks \cite{yang2020hyperparameter}. The simplest one is known as exhaustive search which looks for finding the best parameters by checking as many possible solution as possible based on the level of desirable performance or computational resources. Two well-known subcategory of this type are random search which randomly explores the search space, and grid search which systematically discretizes the search space and checks the solutions that fall on in grid. Other more intelligent methods which are developed for a more effective search are including Gradient based methods, Reinforcement Learning and Evolutionary Algorithms. Gradient based approaches are iterative search algorithms which attempt to find the optimal solutions according to gradient value at each step. These algorithms often gets stuck in local optima. Reinforcement learning algorithms, on the other hand, is a search paradigm that traverses solution space based on taking actions from a predefined list. Each action changes the state of the agent and induces a reward. The goal is to maximize the return or the sum of discounted rewards. Evolutionary optimization is referred to bio-inspired methods in which a population of solutions are created, evolved and evaluated iteratively. Three popular Evolutionary algorithm variants are Genetic Algorithms (GA), Genetic Programming (GP) and Evolutionary Strategies (ES) \cite{chiong2012variants}. Evolutionary algorithms have attracted attention of researcher due to multiple unique properties. They can search the space in a parallel and independent manner,  they are not based on gradient calculation and hence do not suffer from exploding or vanishing gradients \cite{such2017deep}. Due to these intriguing features, in this paper, we have applied a Genetic algorithm  which is described in the next section. We follow the basic execution steps of GA algorithms. Each components of genetic algorithm is problem dependent and needs to be tuned and adapted to the task based on experimentation \cite{back2000emperical} , \cite{eiben1999parameter}. There are different possibilities for defining each GA operator and component. Chromosomes can be encoded in a binary or value representation. Selection algorithm could be roulette wheel, rank selection or tournament. Crossover could be single point, k-point or partially mapped. Mutation could be inversion or reverse \cite{katoch2021review}. The common approach is to adjust all these parameters according to the application in hand. In our experiments, we followed parameter tuning strategy for selection of operators and applied parameter control for crossover and mutation rates. There are several variants of GA algorithms, such as hybrid GA, parallel GA and chaotic GA \cite{katoch2021review}. The objective of this research is not to investigate and compare these variants, but to leverage classic GA as a meta-heuristic optimization method to tune SSL algorithm's operators. We have made the comparison between four SSL algorithm in three different settings of fully supervised, default SSL and improved with GA algorithm.

\begin{table}
\caption{\label{tab:aug_int}All augmentation operators used and the associated intensity ranges. This set of operators is the same as AutoAugment \cite{cubuk2019autoaugment}.}
\begin{tabular}{lll}
\toprule
{} &    \textbf{Augmentation} & \textbf{Intensity Range} \\
\midrule
  &  HorizontalFlip &        0.0, 1.0 \\
  &    VerticalFlip &        0.0, 1.0 \\
  &          ShearX &        0.0, 0.3 \\
  &          ShearY &        0.0, 0.3 \\
  &      TranslateX &           0, 14 \\
  &      TranslateY &           0, 14 \\
  &          Rotate &         -30, 30 \\
  &           Color &        0.1, 1.9 \\
  &        Solarize &        0.0, 1.0 \\
  &        Contrast &        0.1, 1.9 \\
 &       Sharpness &        0.1, 1.9 \\
 &      Brightness &        0.1, 1.9 \\
\bottomrule

\end{tabular}
\end{table}

\section{Implementation Details}
This work is interested in exploring a massive amount of different configurations of augmentations in the pretext task of common SSL algorithms, in order to do so, two standard benchmark which are commonly used for image classification are used. We explore the performance of four SSL algorithms using two different batch sizes of 32 and 256. We carry out the experiments on CIFAR-10 and SVHN datasets. In the following subsections, we provide more in-depth details about the outcome of each experiment.

\subsection{Dataset}
\textbf{CIFAR-10}. The CIFAR-10 dataset \cite{Krizhevsky09learningmultiple} is a popular coloured image dataset consisting of 10 classes of airplane, automobile, bird, cat, deer, dog, frog, horse, ship, and truck. This dataset consists of 60000 32x32 images, with 50000 train images and 10000 test images. The test set consists of 1000 randomly sampled instances from each class, and the remaining 50000 are used for the training set. The classes in this dataset are mutually exclusive and the creators took extra caution to ensure the similar classes of truck and automobile contain no overlap.  

\textbf{SVHN}. The Street View House Numbers (SVHN) Dataset \cite{Netzer_SVHN} is an image dataset consisting of real world images of house numbers taken from Google street view images. This dataset is presented as a more challenging version of MNIST \cite{deng2012mnist} consisting of 600,000 32x32 labeled digits which are cropped from street view images. The classification problem posed in SVHN is significantly harder than MNIST as it contains real-world images of numbers. 
\subsection{Network Architecture and Training Specifications}
\begin{figure}[!h]
\begin{center}
\includegraphics[width=12cm]{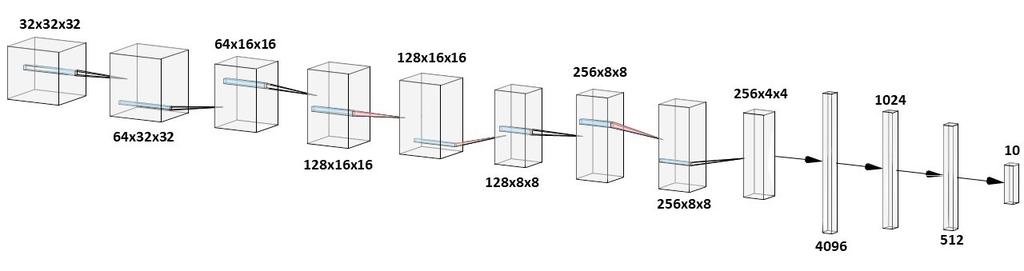}
\end{center}
\caption{\label{fig:backbone} Architecture of Network used for the downstream task.}
\end{figure}

For all experiments, a small and simple convolutional neural network is used as the backbone for both pretraining and downstream tasks. As illustrated in Figure \ref{fig:backbone}, this network consists of 3 convolutional blocks. Each block contains two convolutions with a kernel size of 3x3, followed by ReLU activation functions, a max pool and a batch normalization operation. For each SSL algorithm, i.e., BYOL, NNCLR, SimSiam and SwAV, this backbone is applied in accordance to the specific algorithm. All the hyperparameter used for network training of each of the SSL algorithm are according to the original papers. For the downstream classification task, all experiments use an Adam optimizer with a learning rate of 0.001 and weight decay of 0.0005. Cross entropy is used as loss function.  In addition, a three layer linear model with ReLU activation is used on top of the pretrained networks. We train both the pretext and downstream tasks for 10 epochs. The backbone is first pretrained using one of the four SSL algorithms then the linear head is added to the network for fine-tuning on the downstream task. For each experiment, we roughly exploit 4GB of GPU and 12 CPUs.

The implementation of the proposed work relies on several different libraries. For the deep learning components, such as model architectures, dataset loading, and supervised training, PyTorch is used \cite{torch_paszke}. Building on PyTorch, the python library, Lightly \cite{susmelj2020lightly}, implements cutting-edge SSL algorithms, this library is used for implementing the four SSL algorithms in this research. To implement the evolutionary mechanism in this research, the python library Deap \cite{DEAP_JMLR2012} is employed. 

\section{Method}
We propose two evolutionary mechanism for augmentation operator optimization through a genetic algorithm to find the best performing chromosomes and corresponding models. The first approach is a single optimization method, in which augmentation policy of each one of the four SSL algorithm is optimized individually, whereas the second approach is a multi optimization method through which not only we optimize the augmentation policy but also optimize the downstream task performance by finding the best performing algorithm.
We apply a genetic algorithm to optimize augmentation policies for four different cutting-edge self-supervised algorithms. From a genetic algorithm perspective, the key area that requires the most attention is the definition of the fitness function. The proposed fitness function serves as a proxy function that aims to gauge the impact of changing the augmentation policies for a given self-supervised algorithm. Below, our formulation of the fitness function is elaborated. To represent the augmentation strategies as chromosomes we employ a value representation. We experiment with two different flavors of evolutionary augmentation optimization, namely single optimization (SO) and Multiple Optimization (MO) modes. With SO we choose one specific SSL algorithm and exclusively optimize the augmentation operators which are presented in each gene, while with MO, one gene in the chromosome specifies the SSL algorithm and the best algorithm will be found through optimization process. Hence, the major difference between SO and MO lies within the chromosome structure which is elaborated in detail in section \ref{sec_chrom} .

\subsection{Chromosome Definitions and Augmentation Policies} \label{sec_chrom}

Taking inspiration from AutoAugment and the preceding work in learning augmentation policies, the proposed value encoding assumes a set of $k$ augmentations, $A = \{a_{1}, a_{2}, \dots, a_{k}\}$. For each augmentation $a_{i}$ an intensity value $i$ can be defined. Authors of \cite{cubuk2019autoaugment} defined a range of values for intensity of each augmentation operator. For the purpose of this work, these ranges have been borrowed. Table \ref{tab:aug_int} shows all the different operators used in our experiments along with their possible $i$ values. 
Let $i_k$ be the intensity value assigned to augmentation operator $a_k$ and $K$ be the maximum number of augmentation operators, in SO, we formulate an augmentation policy in a chromosome as $C_{i} = \{(a_{1}, i_{1}), (a_{2}, i_{2}), \dots, (a_{l}, I_{l})\}$, where $l \leq K$. All of the studied algorithms in this paper employ the augmentation strategy used by SimCLR  \cite{chen_simple_2020}. It was found that no single transformation suffices to learn good representations and when composing many augmentations the task becomes harder but the learned representation is improved significantly \cite{chen_simple_2020}. In this study, we use a value of 3 for $l$. Theoretically, it would be possible to obtain further improvement with a higher number of augmentation operators, but it will increase the computational cost of running the experiments. The choice of 3 augmentation operators is used to attempt to strike a balance between the computational burden, simplicity of the problem and the quality of the learned representation. In MO, we represent a chromosome as $C_{i} = (alg_j, \{((a_{1}, i_{1}), (a_{2}, i_{2}), \dots, (a_{l}, i_{l})\})$ where $alg_j$ is one of the SSL algorithms in $\{SwAV,NNCLR,Simsiam,Byol\}$. 

\subsection{Fitness Function} 
The proposed fitness function aims to evaluate augmentation policies for SSL algorithms based on downstream tasks. More specifically, the test accuracy in the downstream supervised task serves as the fitness for augmentation optimization. It is important to note that in this work we are focused on the augmentation policies for the pretext task, and a fixed set of augmentations are used for the supervised downstream task. 
The most crucial part of our fitness function is the fixing of randomness. By fixing every random component of the deep learning pipeline, the difference between two augmentation strategies can be soundly compared in relative terms. In order to compensate for bias due to fixing the random components of the deep learning pipeline we evaluate our method using $N$ different random seeds for the neural network initialization and data shuffling. The seed values for each experiment are shown in Table \ref{tab:num-seeds-total}.

\subsection{Selection}
There exists a large swath of different selection methods such as Roulette, tournament and Rank \cite{o2009riccardo}. Our method opts for a simple and common approach of Roulette. In this approach, every chromosome has an equal chance of being selected proportional to their fitness. 

\subsection{Crossover}
When selecting a crossover operation caution must be taken to prevent the children chromosomes from having duplicate augmentations. To handle this issue, Partially Mapped Crossover (PMX) \cite{reeves_genetic_2003} is employed. This method avoids creation of chromosomes with duplicate genes. When performing crossover between two or more chromosomes, the mechanism for preventing duplicate genes is based purely on the augmentation operator and not the intensity. When two chromosomes are crossed over, the shared genes are completely copied over to maintain the same intensity as the parent's gene. When performing crossover for our MO approach, PMX is applied to the augmentation policy portion of the chromosome, but for the SSL gene, it is probabilistically  swapped with the other chromosome. 

\subsection{Mutation}
For mutation, only the intensities of the augmentation operators are mutated. In order to perform this, a custom mutation function is created. We call this function $MutGaussianChoice$. This mutation operator probabilistically mutates the genes intensity values within the acceptable range of intensity for that specific operator. The intensity is increased or decreased incrementally by the range of the intensity values (as shown in Table \ref{tab:aug_int}) divided by the increment value which is computed using equation \ref{eq:increment}. For MO, $MutGaussianChoice$ is applied to the augmentation policy portion of the chromosome, but for the SSL gene, the gene is randomly mutated to a different SSL algorithm. 
        \begin{align} \label{eq:increment}
           increment  =  \frac{max(i_{range}) - min(i_{range})}{10}
        \end{align}

\subsubsection{Adaptive Mutation Rates}
When applying Mutation and Crossover, whether to apply the operation to the chromosome(s) is probabilistically determined using a mutation and crossover rate. Our method employs adaptive crossover and mutation rates (i.e. $p_c$ and $p_m$) inspired by \cite{srinivas1994adaptive} as shown in Equations \ref{eq:ada_cx} and \ref{eq:ada_mu}:
        \begin{align}\label{eq:ada_cx}
           p_{c}  = (f_{max} - f') / (f_{max} - \bar{f}), f' \geq \bar{f} \nonumber \\
           p_{c} = 1, f' < \bar{f}
        \end{align}
        \begin{align}\label{eq:ada_mu}
           p_{m}  = (f_{max} - f) / (f_{max} - \bar{f}), f' \geq \bar{f} \nonumber \\
           p_{m} = 0.5, f < \bar{f}
        \end{align}

where $f_{max}$ and $\bar{f}$ denote the maximum fitness value and the average fitness value of the population respectively, and $f'$ denotes the larger fitness of the two chromosomes to be crossed or mutated.

\section{Experiments}
In this work, we aim to understand the impact of augmentation operators used in the pretext task of four popular SSL algorithms. By using an evolutionary mechanism to evolve augmentation policies, hundreds of models are trained during one run of the algorithm. To provide baselines for comparison, we train models in both supervised and self-supervised fashions with the augmentation operators according to the original papers. This provides us with an insight into how our evolved augmentation pipelines compare with a purely supervised approach as well as SSL approaches using the original augmentation pipelines. Given that we train the networks under a constrained setting (i.e., using a small network and minimum number of training epochs), we conduct another experiment in which we study how the best evolved augmentation pipelines perform when trained for more epochs in both the pretext task and downstream task. For this experiment, we  exploit the optimal augmentation operators we found through our proposed evolutionary search mechanism.

\textbf{Supervised setting}. For training the supervised baseline models, we conduct the supervised classification task using all of the same hyperparameters as the downstream task in the SSL experiments with no pretraining, i.e., we train a randomly initialized architecture from scratch. This baseline gives us the ability to compare the representations learned with self-supervision and a randomly initialized representation. In all supervised baselines we use the same configuration described in the downstream task. 

\textbf{Self-supervised baselines}.To understand how the learned augmentation pipelines compare with the original SSL algorithm, we train models using the augmentations used in the original papers. Because our specific training configuration is significantly constrained in terms of network architecture and trarining epochs compared to the original papers, these baselines serve as a method for us to understand how the SSL algorithm behaves in our constrained setting. When training these baselines, the exact same hyperparameters such as batch size, and number of epochs as the algorithms trained with evolved augmentations are used. This allows us to understand how changing the augmentations in our setting affect the SSL algorithm with respect to the original augmentation configuration. 

\textbf{Evolutionary SSL}. The fitness function in the proposed GA employs the downstream test accuracy as its metric. The hyperparameterization of the SSL configuration in our evolutionary based search is the exact same as the self-supervised baseline, except for the augmentations in the pretext task. Precautions were taken to ensure that all random components of the SSL pipeline were controlled with a fixed seed, ensuring that the only component of the pipeline that changes is the augmentations itself. Both the default SSL configuration and Evolutionary Optimized Augmentation configuration using the evolved augmentation use the exact same downstream configuraiton as the supervised baseline. Because of this, with the basic SSL and supervised baselines, we are able to compare how the SSL pretraining using an evolved augmentation policy improves the performances achieved by both the supervised and SSL baselines.

 \textbf{Effect of batch size}
 We study the effect of batch size in the pretext task, and so batch sizes of 32 and 256 were experimented for all the supervised, self-supervised and evolutionary self-supervised modes. 

\textbf{Effect of number of epochs} Given that in our experiments are conducted in a constrained setting, the evolutionary algorithm utilized a low number of epochs. To better evaluate the performance of SSL algorithms, we carry out an experiment, in which we use the best found augmentation policies by our evolutionary search mechanism and train SSL models for 50, 100, and 1000 epochs in the pretext task and 50 epochs in the downstream task. 

\begin{figure}
 \begin{tabular}{cc}
  \includegraphics[width=80mm]{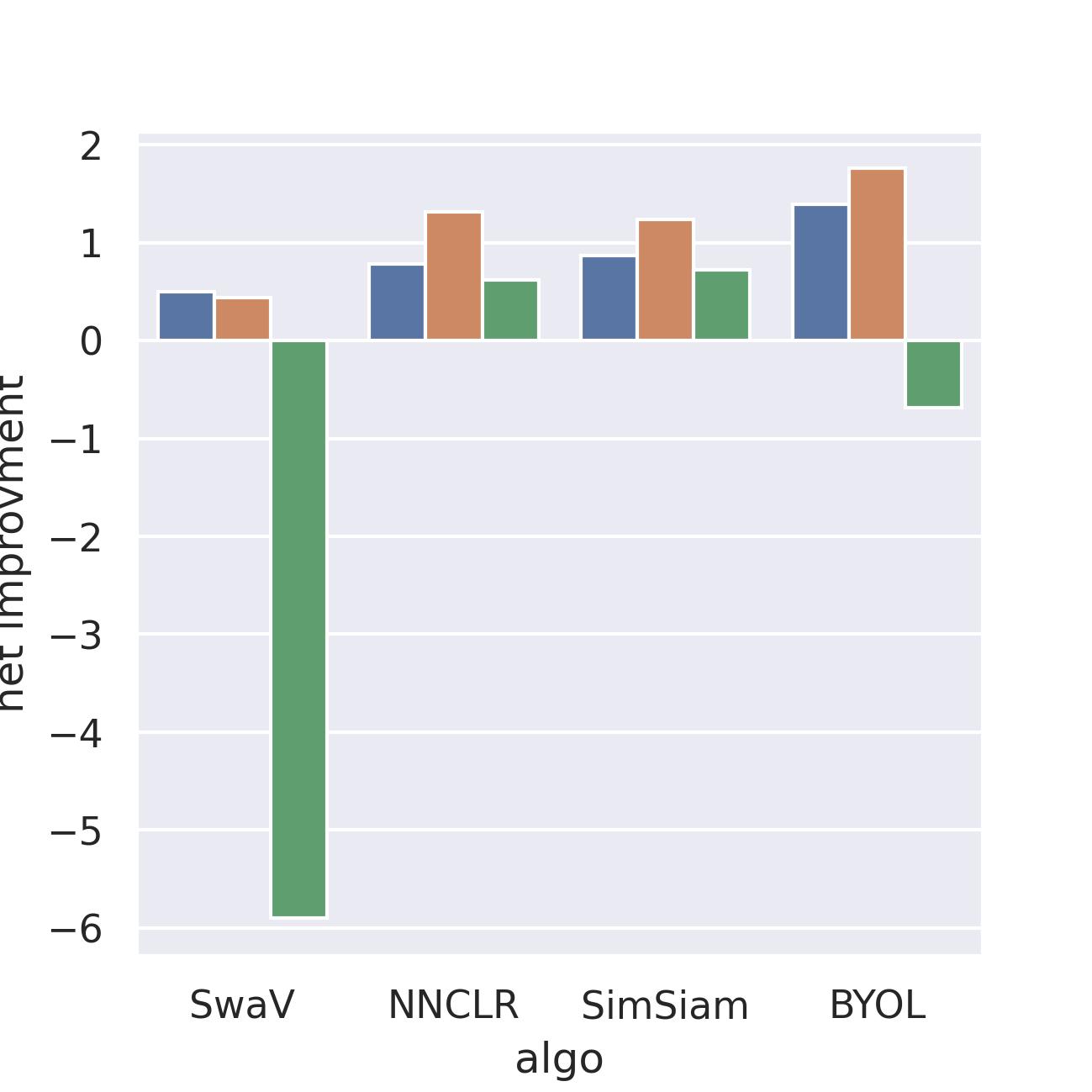} & \includegraphics[width=80mm]{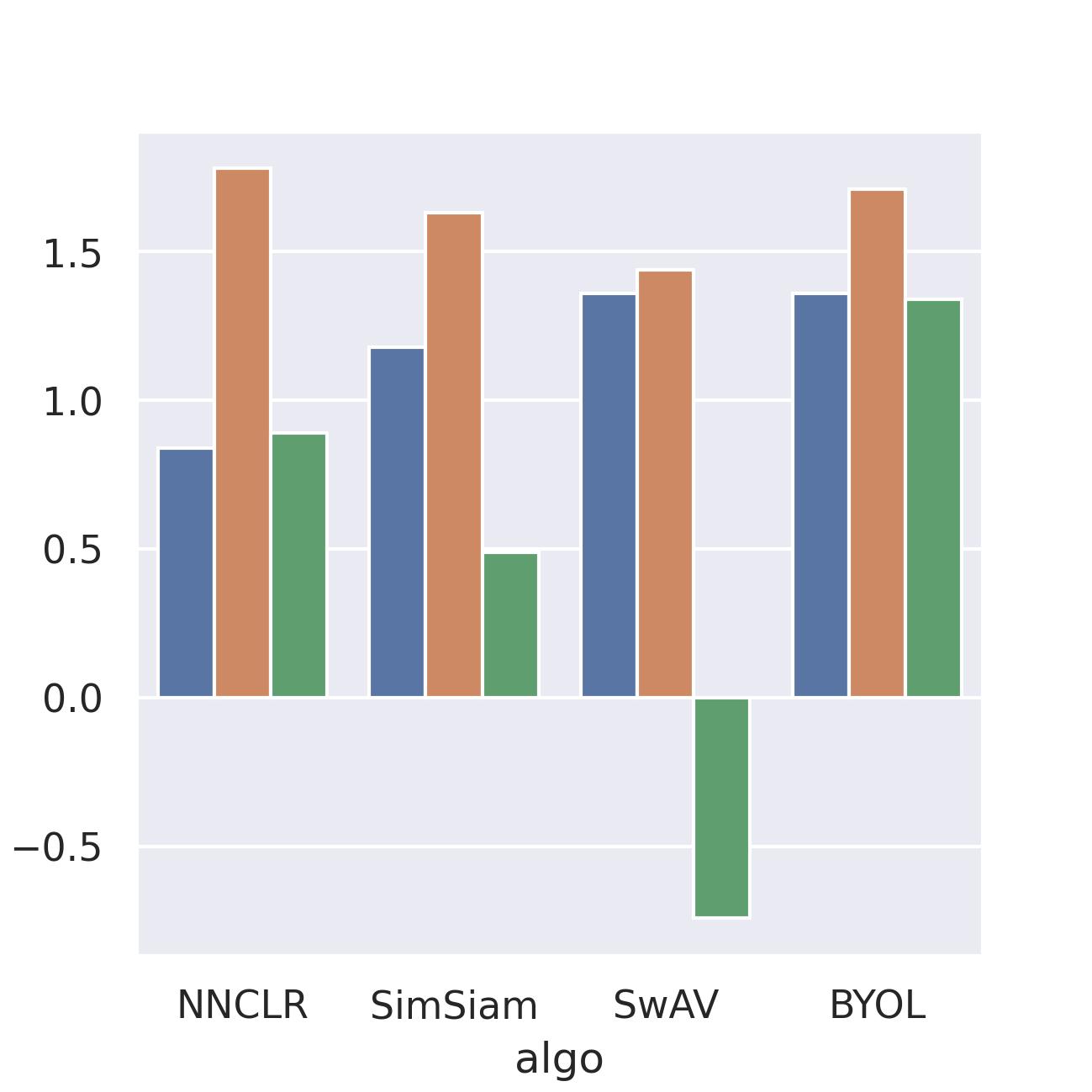} \\
  (a) BS 32, Cifar10  & (b) BS 256, Cifar10  \\
 [6pt]
\end{tabular}
\caption{Training models for more epochs in both the pretext and downstream task. X-axis representing the different SSL algorithms, hue representing the number of epochs used for the pretext task (50, 100 or 1000), and the y-axis representing the improvement on the initial accuracy of the augmentation pipeline.}
\label{fig:train-models-longer-bar}
\end{figure}

\subsection{Results}
\label{sect:results}
As explained in the previous sections, in order to understand the effect of changing the augmentation operators in the pretext task of the four mentioned SSL algorithms, the Evolutionary Algorithm is carried out. For each of the four algorithms, BYOL, SimSiam, SwAV and NNCLR, two datasets of SVHN and Cifar10 and pretext batch sizes of 32 and 256 were used. To  mitigate the bias due to randomness, $N$ random seeds were used to control all randomness in the deep learning components of the algorithm, where $N$ for each experiment is shown in table \ref{tab:num-seeds-total}. This ensures that the differences in training outcomes are not due to randomness in the training pipeline, such as model initialization, or shuffling of data. For each of the SSL algorithms, batch sizes and data sets, our Evolutionary Optimized Augmentation Algorithm was employed with a population size of 15. Running for 10 generations, with a population size of 15 and for the number of seeds shown in Table \ref{tab:num-seeds-total} resulted in a total of \textbf{15000} different augmentation configurations in competition with one another. Note that due to the nature of the GA, with individuals being carried over to the next generation, not all 15000 configurations are unique. 

\begin{figure}
 \begin{tabular}{cc}
  \includegraphics[width=80mm]{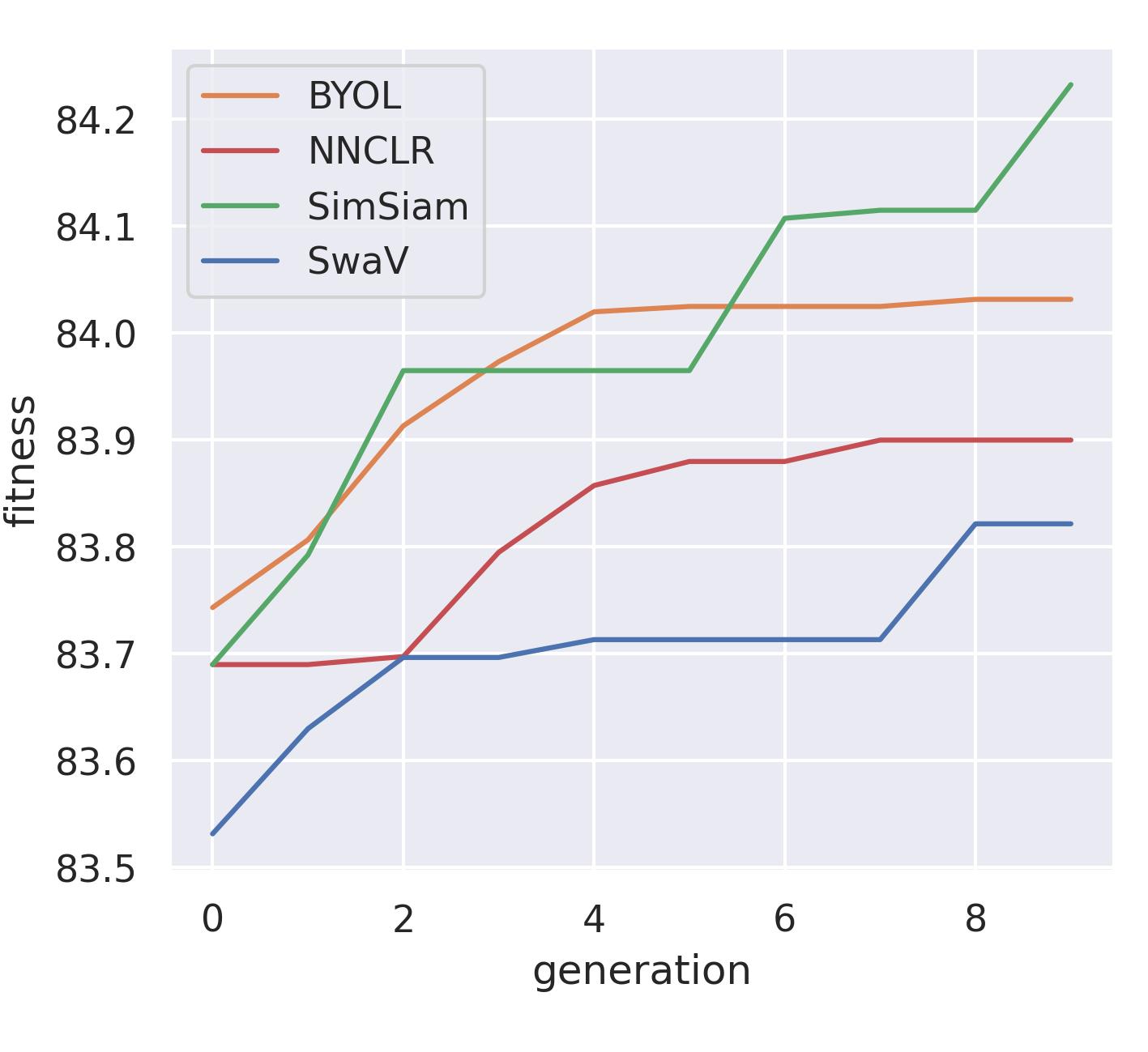} & \includegraphics[width=80mm]{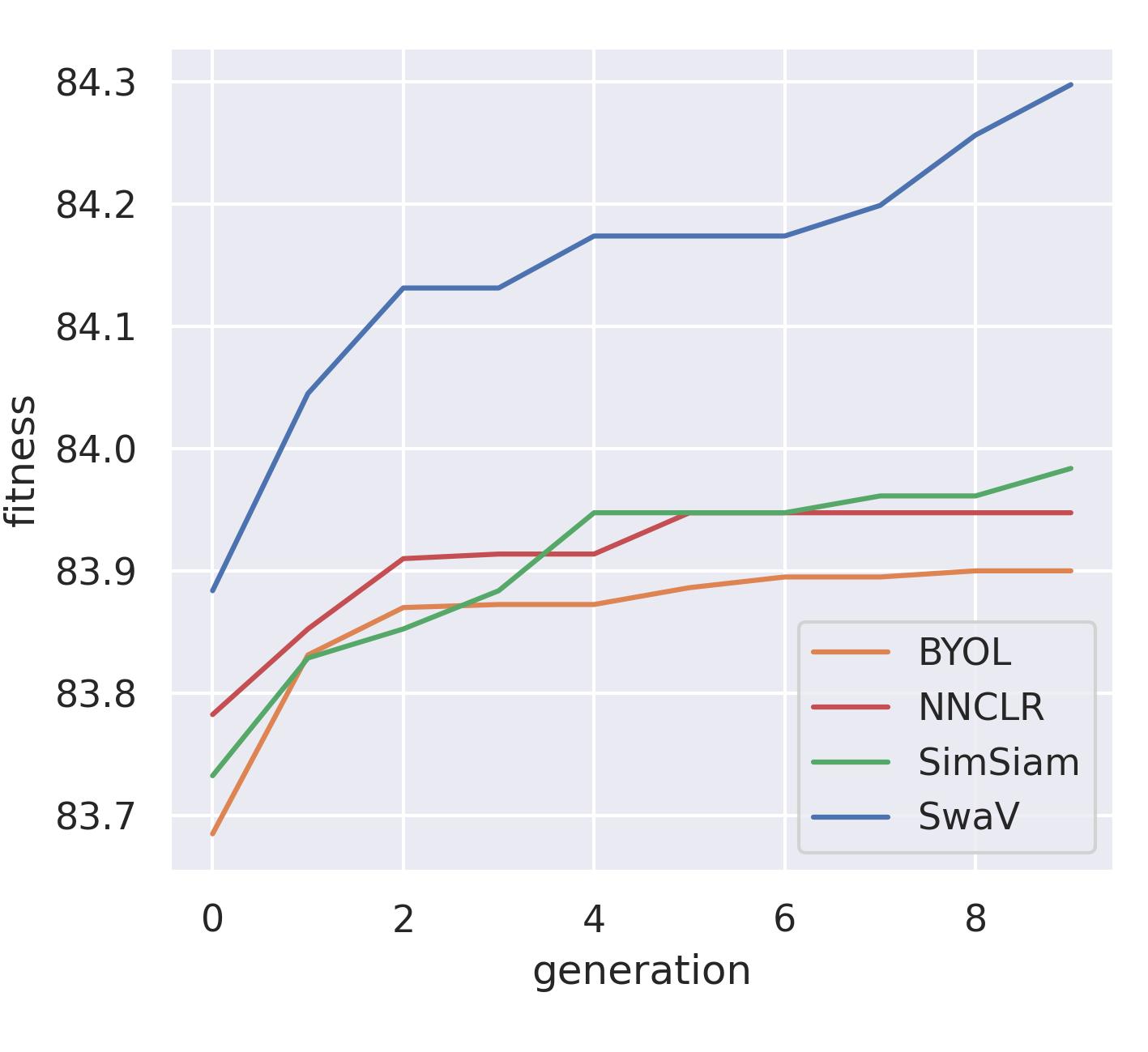} \\
  (a) Cifar10, BS 32  & (b) Cifar10, BS 256  \\
   \includegraphics[width=80mm]{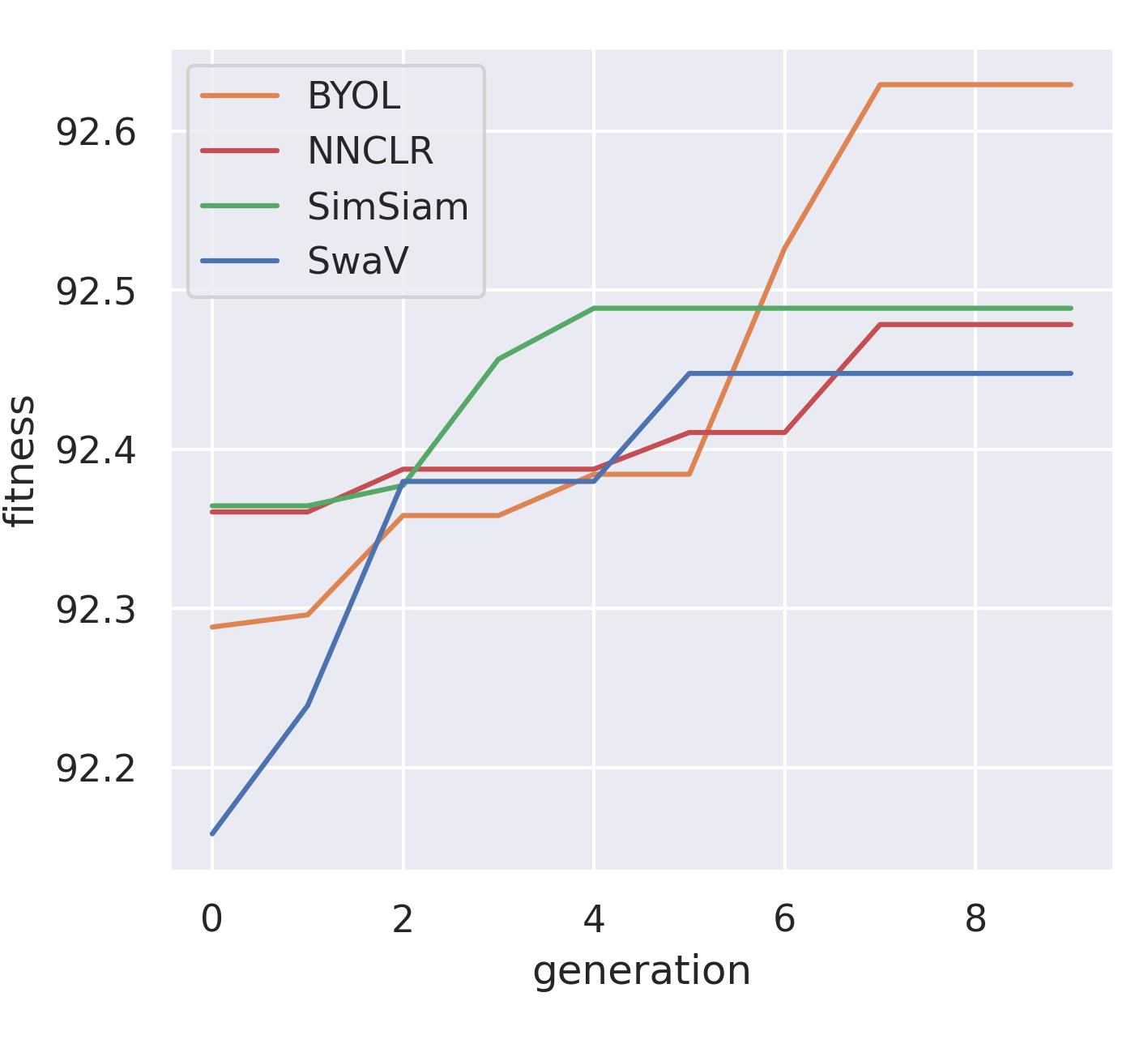} & \includegraphics[width=80mm]{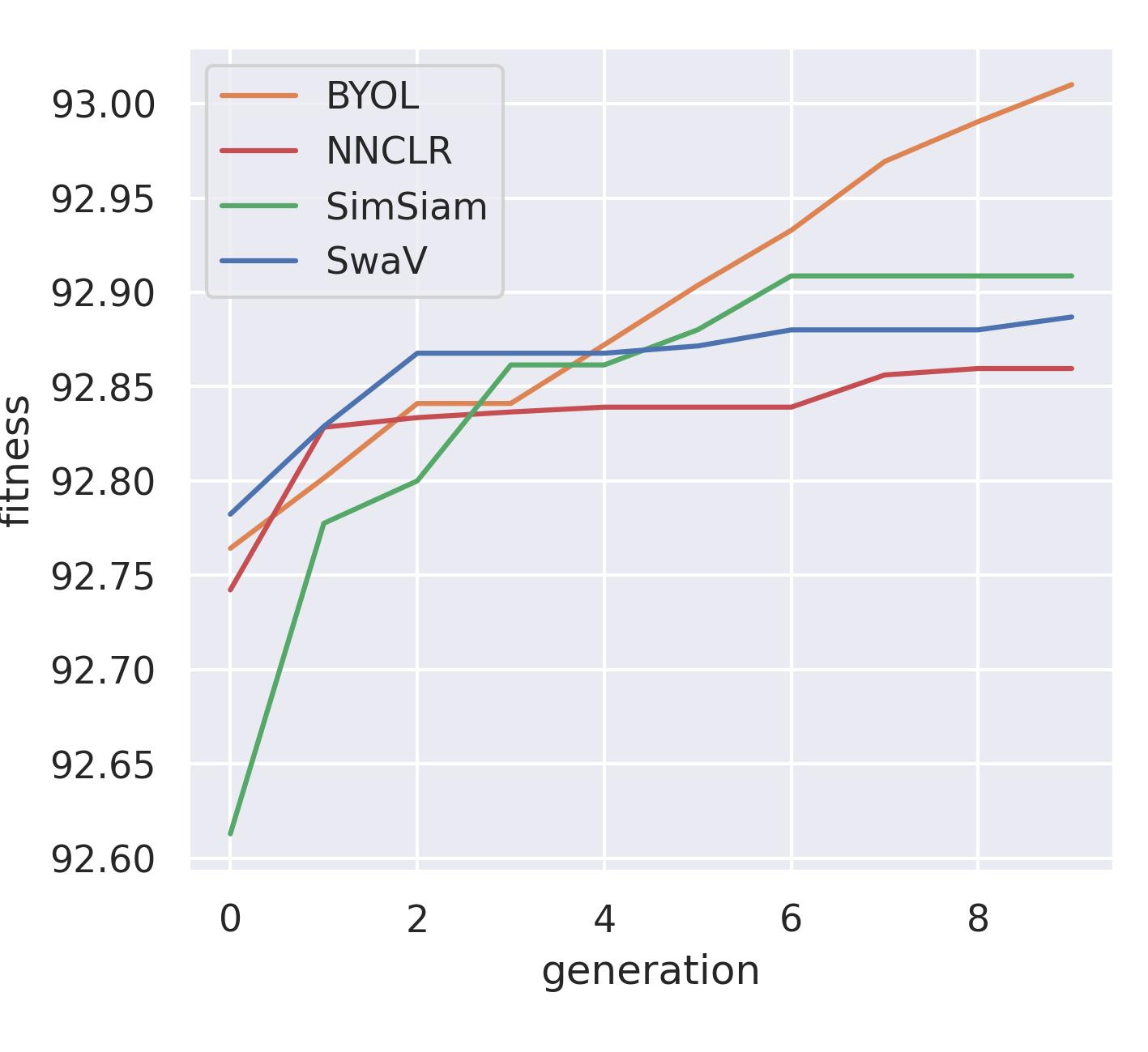} \\
  (c) SVHN, BS 32  & (d) SVHN, BS 256  \\ [6pt]
\end{tabular}
\caption{The average of the best found downstream test accuracy using SO strategy for the supervised classification tasks at each generation for all seeds listed in Table \ref{tab:num-seeds-total}}
\label{fig:GA-avgbest-all}
\end{figure}

\begin{figure}
 \begin{tabular}{cc}
  \includegraphics[width=80mm]{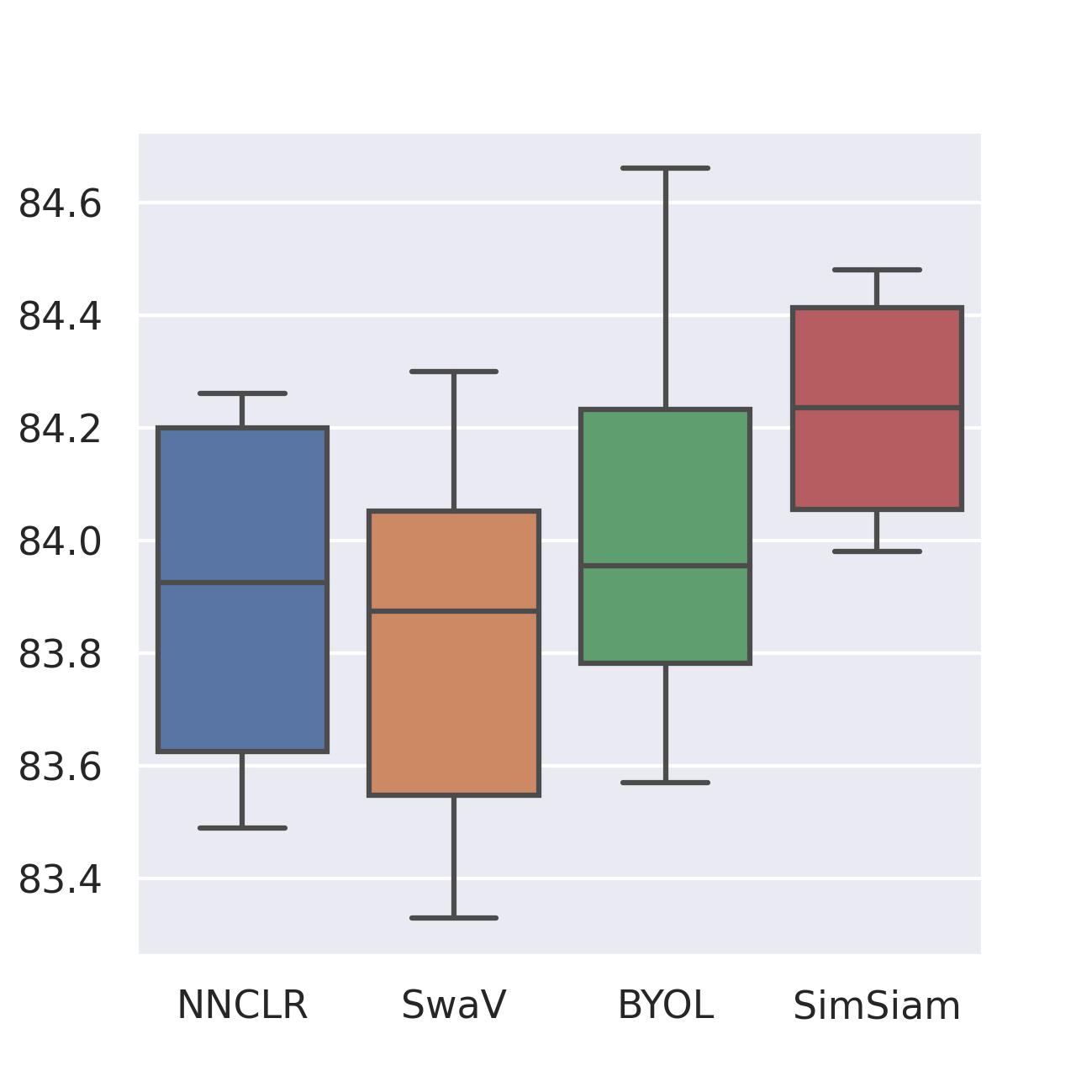} & \includegraphics[width=80mm]{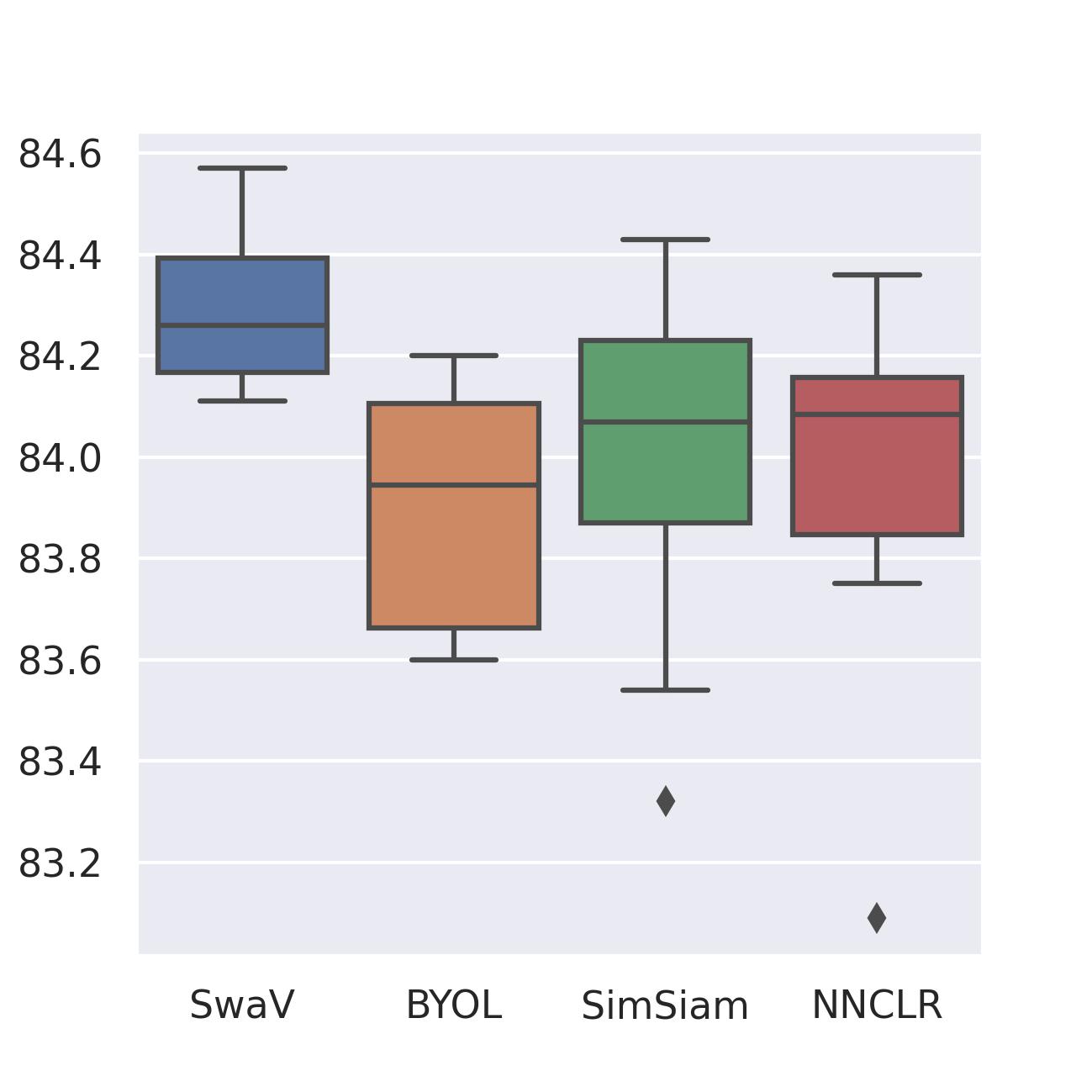} \\
  (a) Cifar10, BS 32  & (b) Cifar10, BS 256  \\
   \includegraphics[width=80mm]{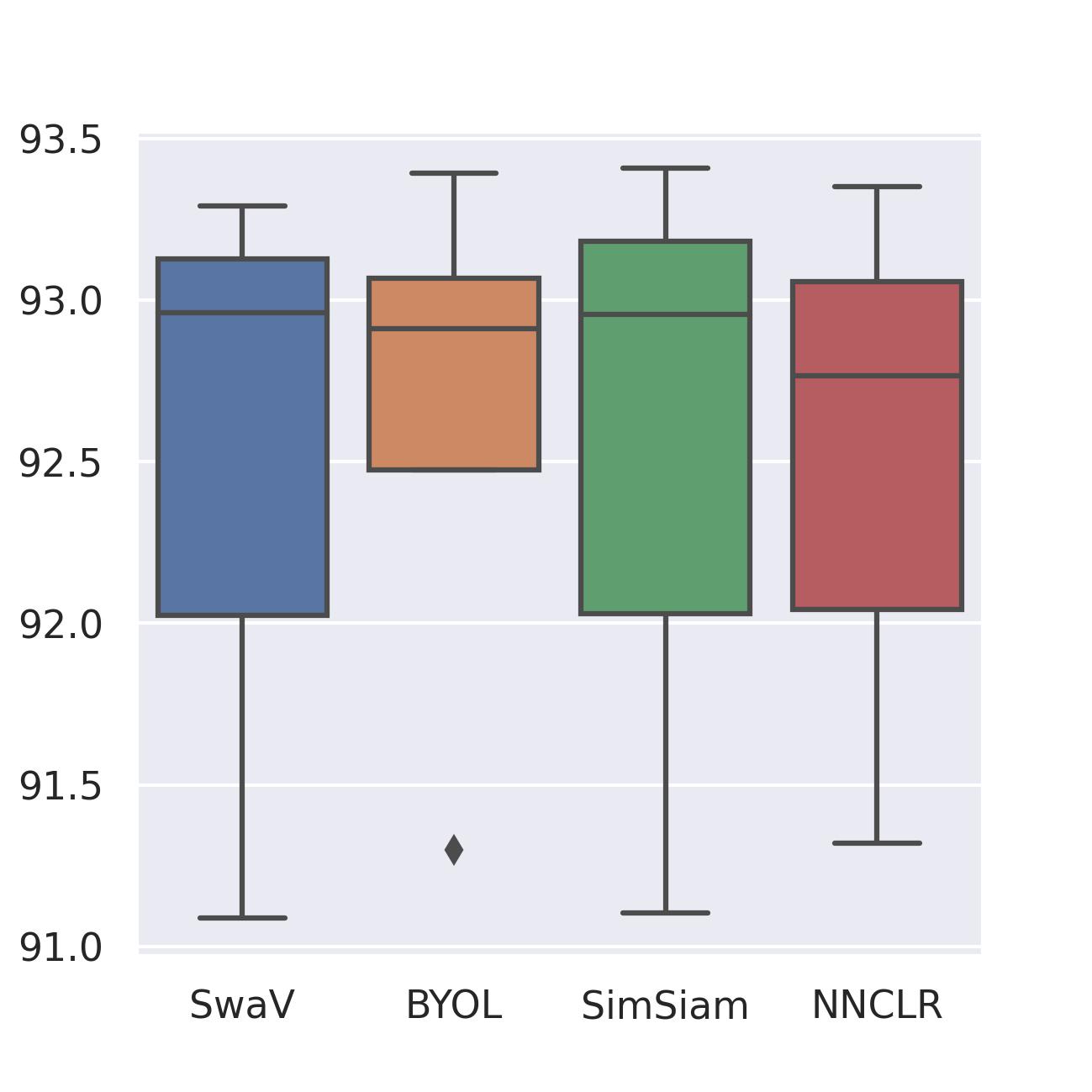} & \includegraphics[width=80mm]{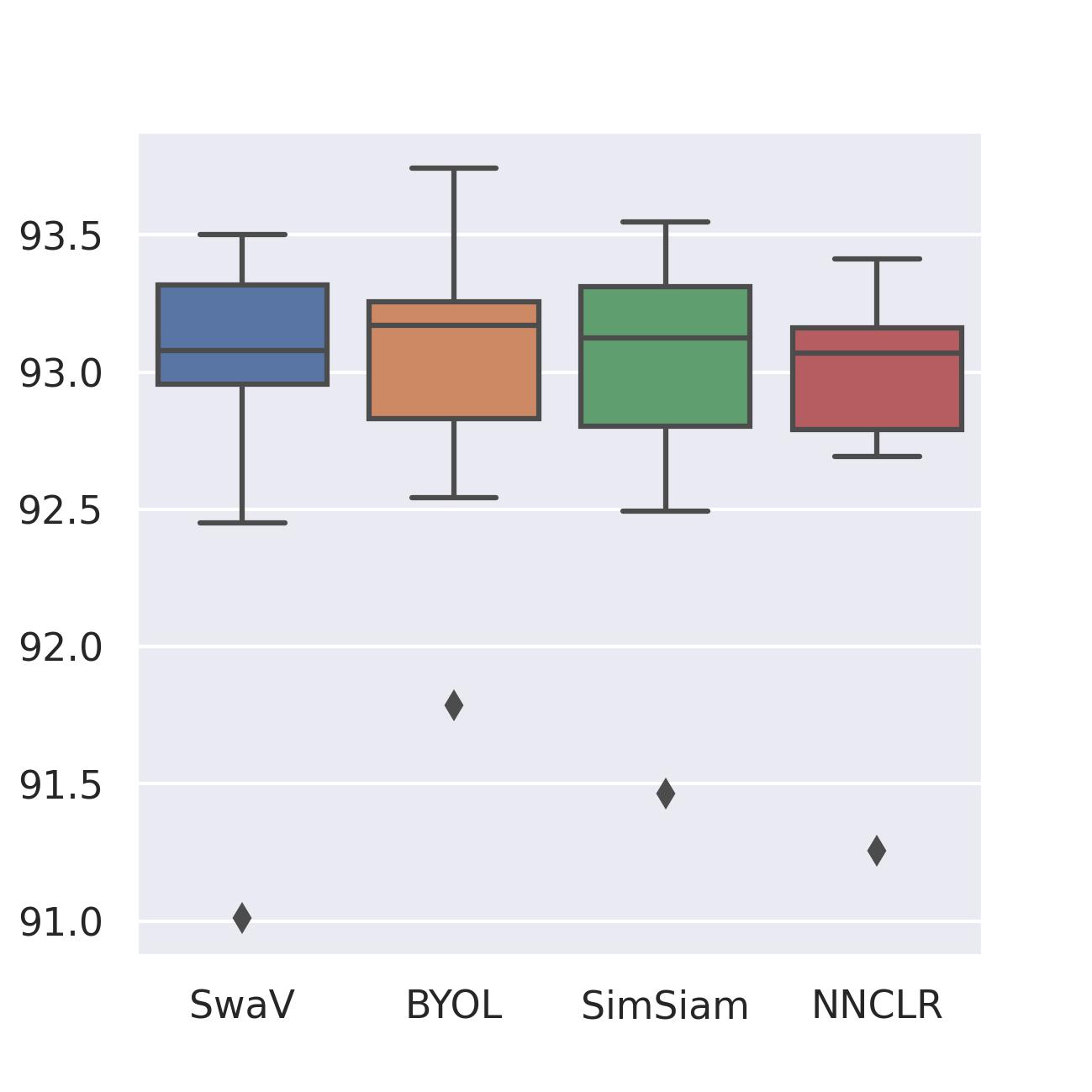} \\
  (c) SVHN, BS 32  & (d) SVHN, BS 256  \\ [6pt]
\end{tabular}
\caption{Comparison of test accuracies of downstream task using SO strategy for different seeds. All used seed values are listed in Table \ref{tab:num-seeds-total}}
\label{fig:GA-avgbest-all-boxplot}
\end{figure}

We find that we can consistently improve the SSL algorithms performance by applying our proposed evolutionary method. In Figure \ref{fig:GA-avgbest-all}, we show the average of the best fitness over all the seeds for each of the four algorithms across generation. More specifically, for each random seed experiment, the best fitness at each generation is extracted and then the average for all seeds is computed. For both Cifar10 and SVHN, an overall monotonic trend of optimization is observed which confirms that our proposed evolutionary method improves the performance of SSL aglorithms. We observe that, mostly, NNCLR accounts for the smallest net improvement, whereas BYOL shows the largest improvement. 
Furthermore, we compare the distribution of classification accuracy of the best solutions of different random initialization found in the final generation of evolutionary process in the both datasets using two different batch sizes in pretext task of SSL. Figure \ref{fig:GA-avgbest-all-boxplot} suggests that the batch size of 256 has a larger impact on the improvement of the accuracy of downstream task classification. In addition, we observe that the results obtained from SVHN dataset presents on par average accuracy for all the four algorithms. In contrast, in CIFAR-10 there is a noticeable variation in the average accuracy of the four algorithms in both batch sizes. With batch size 32, the highest performance is achieved by SimSiam in CIFAR-10 and Byol in SVHN, while with batch size 256, all the four algorithms produce similar results.

\begin{figure}
\begin{center}
\includegraphics[width=12cm]{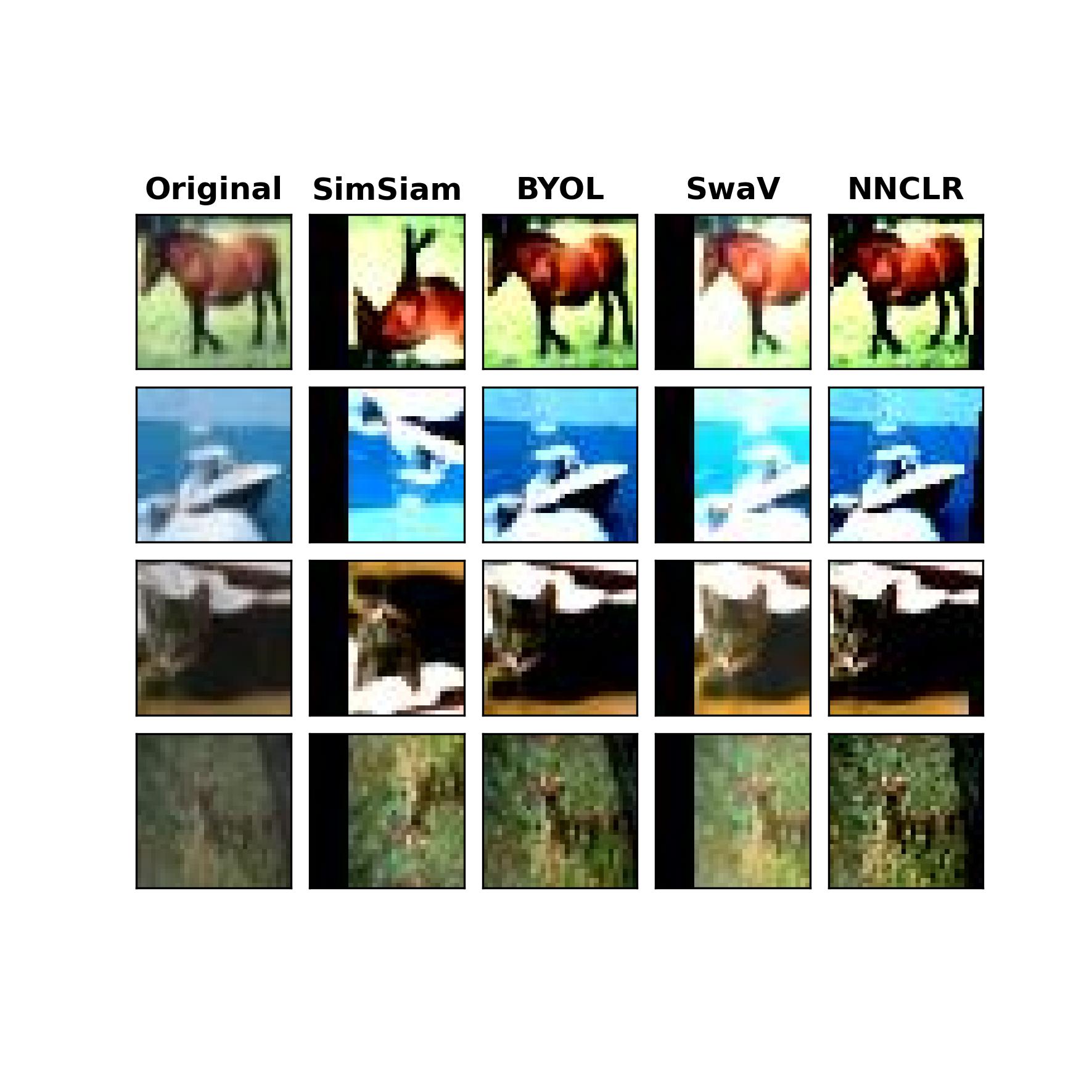}
\end{center}
\caption{\label{fig:best_found_augs_c10}Visualization of best found augmentation policies for cifar10. The policies are presented in Table \ref{table:best_found_augs}. Each row shows the effect of augmentation operator applied to an example image in sequence. }
\end{figure}

\begin{figure}
\begin{center}
\includegraphics[width=12cm]{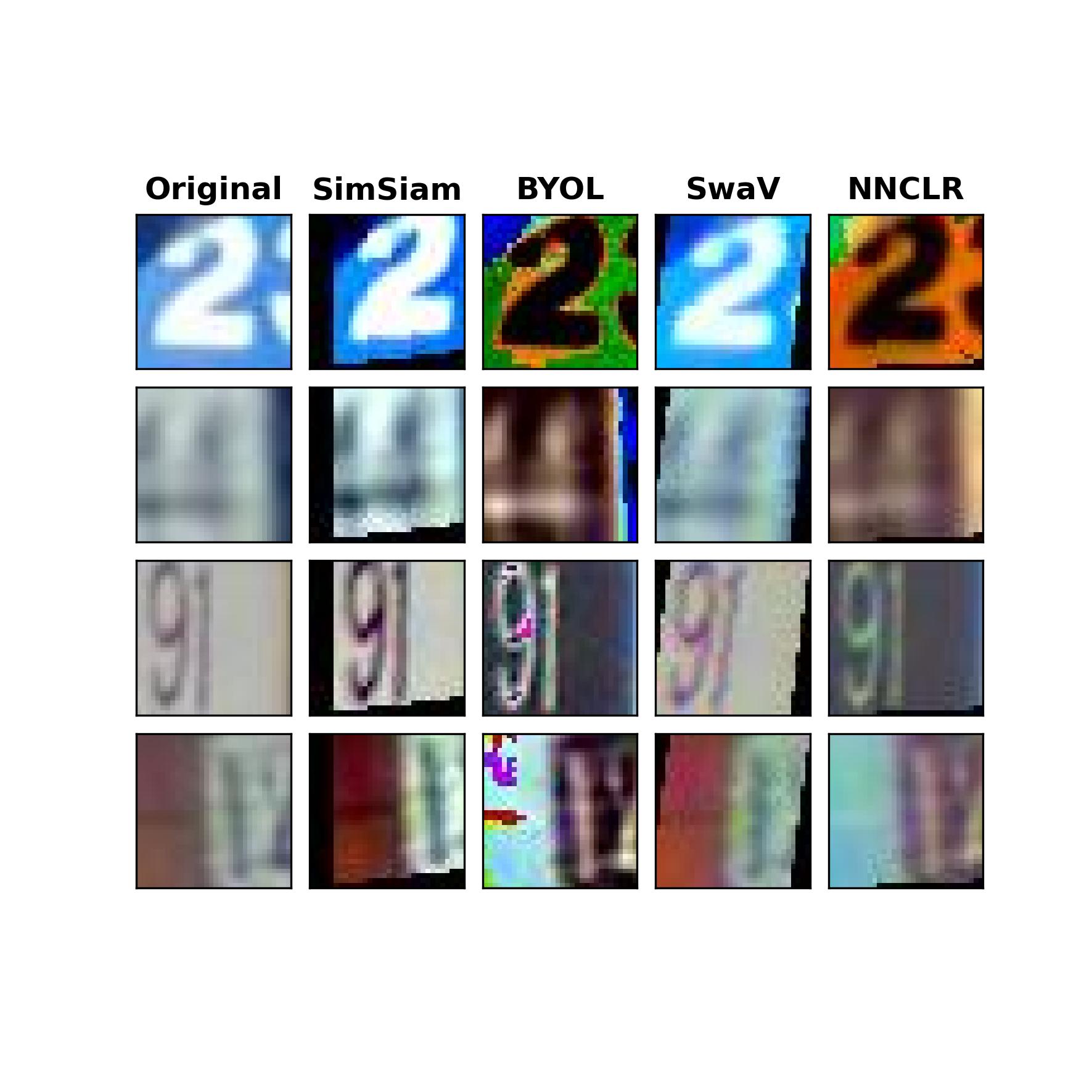}
\end{center}
\caption{\label{fig:best_found_augs_svhn}Visualization of best found augmentation operators for SVHN. The policies are presented in Table \ref{table:best_found_augs}. Each row shows the effect of augmentation operator applied to an example image in sequence.}
\end{figure}

Furthermore, the details of optimal performance obtained by each of the four algorithms are compared in Tables \ref{table:all_seeds_res_c1032}, \ref{table:all_seeds_res_c10256}, \ref{table:all_seeds_res_sv32} and \ref{table:all_seeds_res_sv256}. The results indicate that mostly higher accuracies are achieved when using larger batch size in both datasets. In addition, the solutions found by both SO and MO, outperform those obtained by supervised and self supervised learning.  
In all cases, including both datasets, both batch sizes and for the four SSL algorithms, the SO algorithm finds better solutions than MO. This is due to the fact that single optimization procedure is focused on solutions from a specific SSL method and has a more limited search space. 
More interestingly, in all cases, when we deploy the best configurations found by augmentation operators from SO and train for 50 in pretext and 100 in downstream tasks, we outperform all the results when compared to the supervised training. These results are denoted by SSL(optimized) in the Tables.
In order to ensure that our proposed strategy is consistently achieving better results than the baselines, statistical t-tests were run with acceptance values of 0.95 and 0.99 for comparison of our single optimization strategy with self-supervised baselines presented in Table \ref{table:ttest_ssl}. As evident from the results, it can be observed that the out-performance of the self-supervised baselines is statistically significant in all cases, confirming that our proposed augmentation policy optimization strategy can lead to statistically significant improvement in the accuracy of classification results with SSL algorithms. 
As mentioned before, in all the experiments, we used 10 epochs for both the pretext and downstream tasks. However, in order to better understand how well the proposed evolved augmentation pipelines behave for larger number of epochs in our experiment setting, we conduct another experiment in which we measure the effect of number of epochs on the results. For this experiment, the best augmentation pipeline for each SSL algorithm is used in their respective pretext tasks and are trained for 50, 100, and 1000 epochs. The obtained pretrained models were then fine tuned in the downstream task for 50 epochs. With this experiment, we compare the models being trained for 10 epochs in both the pretext and downstream phases with models that are trained for 50, 100, and 1000 epochs in the pretext task and 50 epochs in the downstream task. As illustrated in Figure \ref{fig:train-models-longer-bar}, we see that there is an increasing boost in performance in downstream task when using 50 and 100 epochs in the pretext task, but this improvement degrades when using 1000 epochs. This finding indicates that training the pretext task for too long results in overfitting and a decay in classification accuracy. 
Furthermore, we summarize the best found policies in Table \ref{table:best_found_augs} and visualize the effect of these policies by a few examples in Figures \ref{fig:best_found_augs_c10} and \ref{fig:best_found_augs_svhn} for CIFAR-10 and SVHN datsets respectively. Details about number of seeds for each experiment is listed in Table \ref{tab:num-seeds-total}.

\begin{table}
\caption{\label{table:best_found_augs}Best found augmentation policies. $aug_i$ refers to the order of augmentation application. $op_i$ refers to the corresponding intensity value}
\begin{tabular}{llllrlrlr}
\toprule
Dataset & ssl algorithm &        $aug_1$ &   $op_1$ &          $aug_2$ &   $op_2$ &      $aug_3$ &   $op_3$ \\
\midrule
svhn &          SwaV &       Color &  1.48 &    TranslateX &  3.00 &    ShearX &  0.13 \\
svhn &          BYOL &   Sharpness &  0.95 &      Contrast &  1.28 &  Solarize &  0.32 \\
svhn &       SimSiam &    Contrast &  1.15 &    TranslateX &  5.00 &    ShearY &  0.12 \\
svhn &         NNCLR &       Color &  0.90 &        ShearY &  0.05 &  Solarize &  0.32 \\
cifar10 &         NNCLR &   Sharpness &  0.88 &      Contrast &  1.18 &    ShearX &  0.10 \\
cifar10 &          SwaV &  TranslateX &  8.00 &    Brightness &  0.69 &     Color &  0.50 \\
cifar10 &          BYOL &    Contrast &  0.97 &     Sharpness &  0.34 &    Rotate & -1.00 \\
cifar10 &       SimSiam &  TranslateX &  8.00 &  VerticalFlip &  0.64 &  Contrast &  1.24 \\
\bottomrule
\end{tabular}
\end{table}

\begin{table}
\caption{\label{tab:num-seeds-total} Seed values for each experiment including pretext batch sizes and datasets. Algo indicates the SSL algorithms optimized with SO strategy.}
\begin{tabular}{llll}
\toprule
exp &     algo &                            seed \\
\midrule
bsize=256, data=svhn &     SwaV &     [0, 1, 2, 3, 4, 5, 6, 7, 8] \\
bsize=256, data=svhn &     BYOL &  [0, 1, 2, 3, 4, 5, 6, 7, 8, 9] \\
bsize=256, data=svhn &  SimSiam &           [0, 1, 2, 5, 6, 7, 8] \\
bsize=256, data=svhn &    NNCLR &     [0, 1, 2, 3, 4, 5, 6, 7, 8] \\
 bsize=32, data=svhn &     SwaV &                       [0, 1, 2] \\
bsize=32, data=svhn &     BYOL &                    [0, 1, 2, 3] \\
bsize=32, data=svhn &  SimSiam &                       [0, 1, 2] \\
 bsize=32, data=svhn &    NNCLR &                       [0, 1, 2] \\
bsize=32, data=cifar10 &    NNCLR &                    [0, 1, 2, 3] \\
bsize=32, data=cifar10 &     SwaV &              [0, 1, 2, 3, 4, 5] \\
bsize=32, data=cifar10 &     BYOL &              [0, 1, 2, 3, 4, 8] \\
bsize=32, data=cifar10 &  SimSiam &                    [0, 1, 2, 3] \\
bsize=256, data=cifar10 &     SwaV &        [0, 1, 2, 3, 4, 5, 6, 7] \\
bsize=256, data=cifar10 &     BYOL &        [0, 1, 2, 3, 4, 5, 6, 7] \\
bsize=256, data=cifar10 &  SimSiam &        [0, 1, 2, 3, 4, 5, 6, 7] \\
bsize=256, data=cifar10 &    NNCLR &        [0, 1, 2, 3, 4, 5, 6, 7] \\
\bottomrule
\end{tabular}
\end{table}

\section{Explainability} 
In this section, we propose two algorithms for analyzing the best models encoded in the optimal performing chromosomes and explain the effect of augmentation operators. 
In the following section, we introduce an augmentation sensitivity and augmentation importance analysis in order to understand the most influential operators for each dataset and SSL algorithms.

\subsection{Augmentation Sensitivity}
 In order to measure the sensitivity of each algorithm to each augmentation operator, we design an ablation study in which we measure the amount of difference in performance in absence of each operator. More specifically, in order to measure the sensitivity of $a_i$ operator, we remove this operator from all chromosomes that include it and replace it with all possible operators from our augmentation set and obtain a new set of chromosomes. Then we obtain the difference between the average performance of this set and the performance of the original chromosome:
OS($a_i$) = performance($a_i$) - average(performance($a_k$)) for all k 
The pseudocode is presented in Algorithm \ref{alg_sens}

\begin{algorithm}
\caption{Computing Sensitivity}\label{alg:cap}
\begin{algorithmic}
\State $op \gets $Operator of interest
\State $C \gets $Set of all Chromosomes
\State $Cop \gets $Subset of chromosomes in $C$ that contain operator $op$
\State $Sensitivity \gets 0$
\For{$c_{i}$ in Cop}
    \State $AvgSimAcc \gets 0$
    \State $TotalSimilarChromos \gets 0$
    \For{$c_{j}$ in C}
        \State $NumEqualOps \gets 0$
        \For{$op_{i}$ in $c_{i}$.operations}
            \State $NumEqualOps \gets NumEqualOps+1$
        \EndFor
        \If{$op$ not in $c_{j}$.operations and $NumEqualOps \equiv c_{i}$.length $- 1$}
            \State $AvgSimAcc \gets AvgSimAcc + c_{j}$.accuracy
            \State $TotalSimilarChromos \gets TotalSimilarChromos + 1$
        \EndIf
    \State $AvgSimAcc \gets AvgSimAcc \div TotalSimilarChromos$
    \State $Sensitivity \gets Sensitivity + |c_{i}$.accuracy $- AvgSimAcc|$
    \EndFor
\EndFor
\State $Sensitivity \gets Sensitivity \div Cop$.length
\end{algorithmic}
\label{alg_sens}
\end{algorithm}

\begin{table}
\centering
\caption{\label{table:all_seeds_res_c1032} Best found accuracy on downstream task for Cifar10 with a batch size of 32 in the pretext task. SO: Single Optimization. MO: Multiple Optimization. SSL (optimized): the best outcomes when training the best-found configurations for longer epochs. The bold italicized values represent the best-found results from the SO, MO and two baselines. The bolded values show the improved results with longer training epochs for the best found SO methods. }
\begin{tabular}{lrrrr}
\toprule
algo &  NNCLR &  SimSiam &   SwaV &   BYOL \\
\midrule
SSL (SO)        &  \textbf{\textit{84.26 }} &  \textbf{\textit{84.48}} & \textbf{\textit{84.30}} &  \textbf{\textit{84.27}} \\
SSL (MO)        &  83.94 &    84.18 &  84.18 &  83.92 \\
SSL (default)   &  82.81 &    83.59 &  81.59 &  83.83 \\
Supervised      &  83.24 &    83.24 &  83.24 &  83.24 \\ \hdashline
SSL (optimized) &  \textbf{85.58} &    \textbf{85.72} &  \textbf{84.74} &  \textbf{86.03} \\
\bottomrule
\end{tabular}
\end{table}

\begin{table}
\caption{\label{table:all_seeds_res_c10256} Best found accuracy on downstream task for Cifar10 with a batch size of 256 in the pretext task. SO: Single Optimization. MO: Multiple Optimization. SSL (optimized): the best outcomes when training the best-found configurations for longer epochs. The bold italicized values represent the best-found results from the SO, MO and two baselines. The bolded values show the improved results with longer training epochs for the best found SO methods.}
\centering
\begin{tabular}{lrrrr}
\toprule
algo &  NNCLR &  SimSiam &   SwaV &   BYOL \\
\midrule
SSL (SO)        &  \textbf{\textit{84.36}} &     \textbf{\textit{84.43}} &  \textbf{\textit{84.57}} &  \textbf{\textit{84.20}} \\
SSL (MO)        &  84.24 &    83.85 &  83.97 &  84.20 \\
SSL (default)   &  83.86 &    83.23 &  83.22 &  84.01 \\
Supervised      &  83.24 &    83.24 &  83.24 &  83.24 \\ \hdashline 
SSL (optimized) &  \textbf{85.90} &    \textbf{86.06} &  \textbf{85.93} &  \textbf{85.81} \\
\bottomrule
\end{tabular}
\end{table}

\begin{table}
\centering
\caption{\label{table:all_seeds_res_sv32} Best found accuracy on downstream task for SVHN with a batch size of 32 in the pretext task. SO: Single Optimization. MO: Multiple Optimization. SSL (optimized): best outcomes when training the best-found configurations for longer epochs.The bold italicized values represent the best-found results from the SO, MO and two baselines. The bolded values show the improved results with longer training epochs for the best found SO methods. }
\begin{tabular}{lrrrr}
\toprule
algo &  NNCLR &  SimSiam &   SwaV &   BYOL \\
\midrule
SSL (SO)        &  \textbf{\textit{93.35}} &    \textbf{\textit{93.41}} &  \textbf{\textit{93.29}} &  \textbf{\textit{93.39}} \\
SSL (MO)        &  93.07 &    93.07 &  93.07 &  92.70 \\
SSL (default)   &  92.34 &    92.52 &  91.88 &  92.85 \\
Supervised      &  91.28 &    91.28 &  91.28 &  91.28 \\ \hdashline
SSL (optimized) &  \textbf{93.67} &    \textbf{93.75} &  \textbf{93.58} &  \textbf{93.55} \\
\bottomrule
\end{tabular}
\end{table}

\begin{table}
\centering

\captionsetup{justification=centering}
\caption{\label{table:all_seeds_res_sv256} Best found accuracy on downstream task for SVHN with a batch size of 256 in the pretext task. SO: Single Optimization. MO: Multiple Optimization. SSL (optimized): best outcomes when training the best-found configurations for longer epochs. The bold italicized values represent the best-found results from the SO, MO and two baselines. The bolded values show the improved results with longer training epochs for the best found SO methods.}
\begin{tabular}{lrrrr}
\toprule
algo &  NNCLR &  SimSiam &   SwaV &   BYOL \\
\midrule
SSL (SO)        &  \textbf{\textit{93.41}} &    \textbf{\textit{93.55}} &  \textbf{\textit{93.50}} &  \textbf{\textit{93.37}} \\
SSL (MO)        &  92.94 &    92.94 &  92.83 &  92.94 \\
SSL (default)   &  92.75 &    91.87 &  91.95 &  92.34 \\
Supervised      &  83.24 &    83.24 &  83.24 &  83.24 \\ \hdashline
SSL (optimized) &  \textbf{93.70} &    \textbf{93.68} &  \textbf{93.63} &  \textbf{93.67} \\
\bottomrule
\end{tabular}
\end{table}

\begin{table}
\centering
\caption{\label{table:ttest_ssl} T-Test results for all Single-Objective experiments against SSL baseline.}
\begin{tabular}{lrllrll}
\toprule
  batch size &  dataset &     algo &         p-val &  95\% C.I. &  99\% C.I. \\
\midrule
         32 &  cifar10 &     BYOL &  4.926368e-07 &      True &      True \\
          32 &  cifar10 &    NNCLR &  2.658124e-05 &      True &      True \\
          32 &  cifar10 &  SimSiam &  2.917474e-06 &      True &      True \\
         32 &  cifar10 &     SwaV &  2.240725e-06 &      True &      True \\
         32 &     svhn &     BYOL &  1.658260e-05 &      True &      True \\
          32 &     svhn &    NNCLR &  1.515886e-04 &      True &      True \\
         32 &     svhn &  SimSiam &  4.092746e-04 &      True &      True \\
         32 &     svhn &     SwaV &  4.330569e-04 &      True &      True \\
        256 &  cifar10 &     BYOL &  1.990317e-12 &      True &      True \\
        256 &  cifar10 &    NNCLR &  3.736046e-12 &      True &      True \\
        256 &  cifar10 &  SimSiam &  1.584149e-12 &      True &      True \\
        256 &  cifar10 &     SwaV &  1.235250e-15 &      True &      True \\
        256 &     svhn &     BYOL &  5.714354e-25 &      True &      True \\
        256 &     svhn &    NNCLR &  5.901998e-21 &      True &      True \\
        256 &     svhn &  SimSiam &  3.431059e-19 &      True &      True \\
       256 &     svhn &     SwaV &  1.401669e-18 &      True &      True \\
\bottomrule
\end{tabular}
\end{table}

\subsection{Augmentation Importance}
In this analysis, for each operator and each SSL method, we calculate the number of times an operator has appeared in the top 50 chromosomes from all the generations (based on test accuracies). We normalized this value by number of operators and number of chromosomes. The details of this procedure are illustrated in Algorithm \ref{alg_imp}
This method allows us to find the degree of utilization of augmentation operators for generating the best found chromosomes. This metric allows us to understand how different augmentations are impacting the top chromosomes. 

\begin{algorithm}
\caption{\label{alg_imp}Computing Operator Importance}
\begin{algorithmic}
\State $op \gets $Operator of interest
\State $C \gets $Set of all Chromosomes
\State $N \gets $ Number of chromosomes to consider
\State $Importance \gets 0$
\State $C \gets$ sorted($C$)
\For{$c_{i}$ in C$[:N]$}
    \If{$op$ in $c_{i}$.operations}
            \State $Importance \gets Importance+1$
    \EndIf
\EndFor
\end{algorithmic}
\end{algorithm}

\subsection{Explainability Findings}
\begin{figure}
 \begin{tabular}{cc}
  \includegraphics[width=80mm]{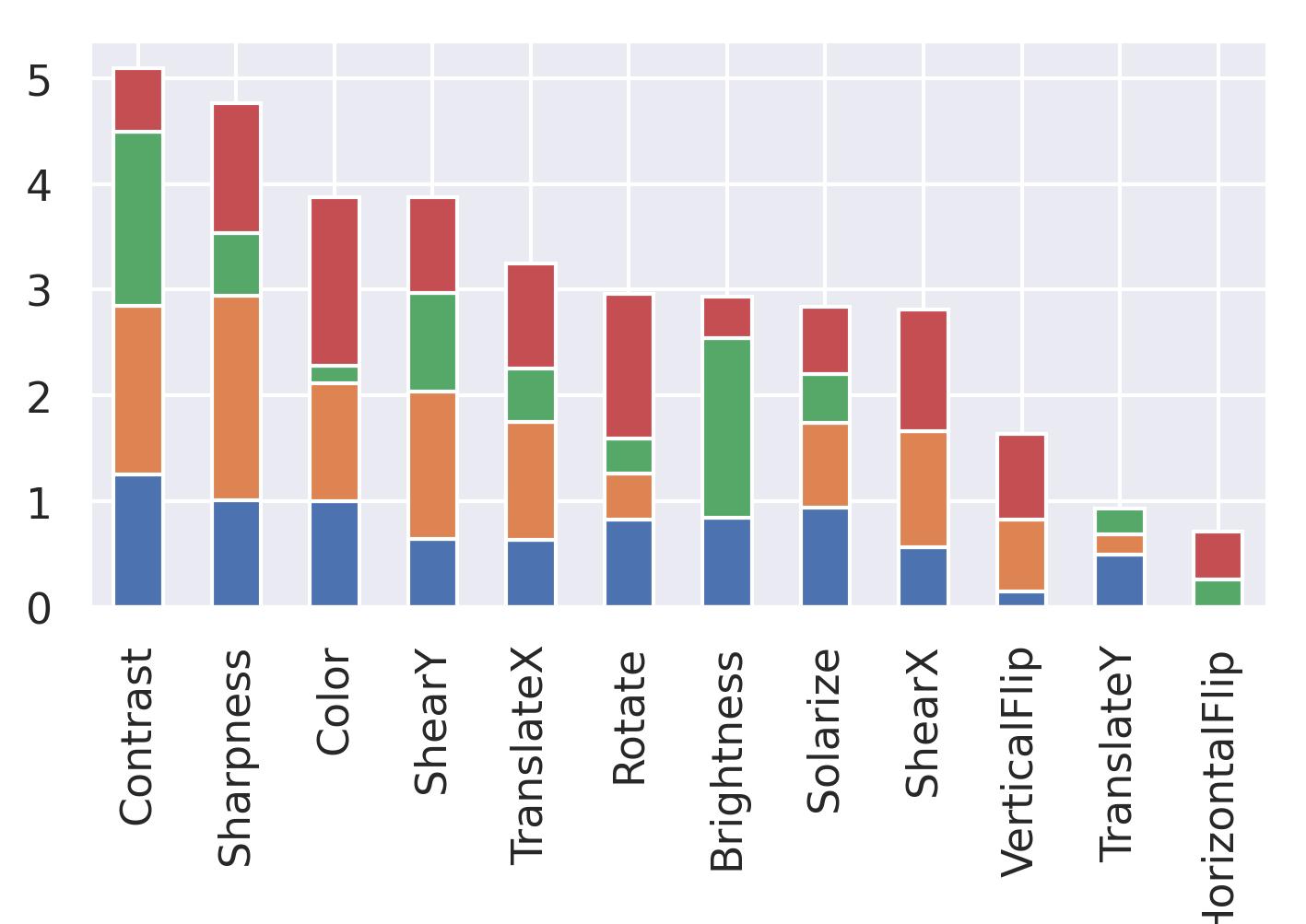} & \includegraphics[width=80mm]{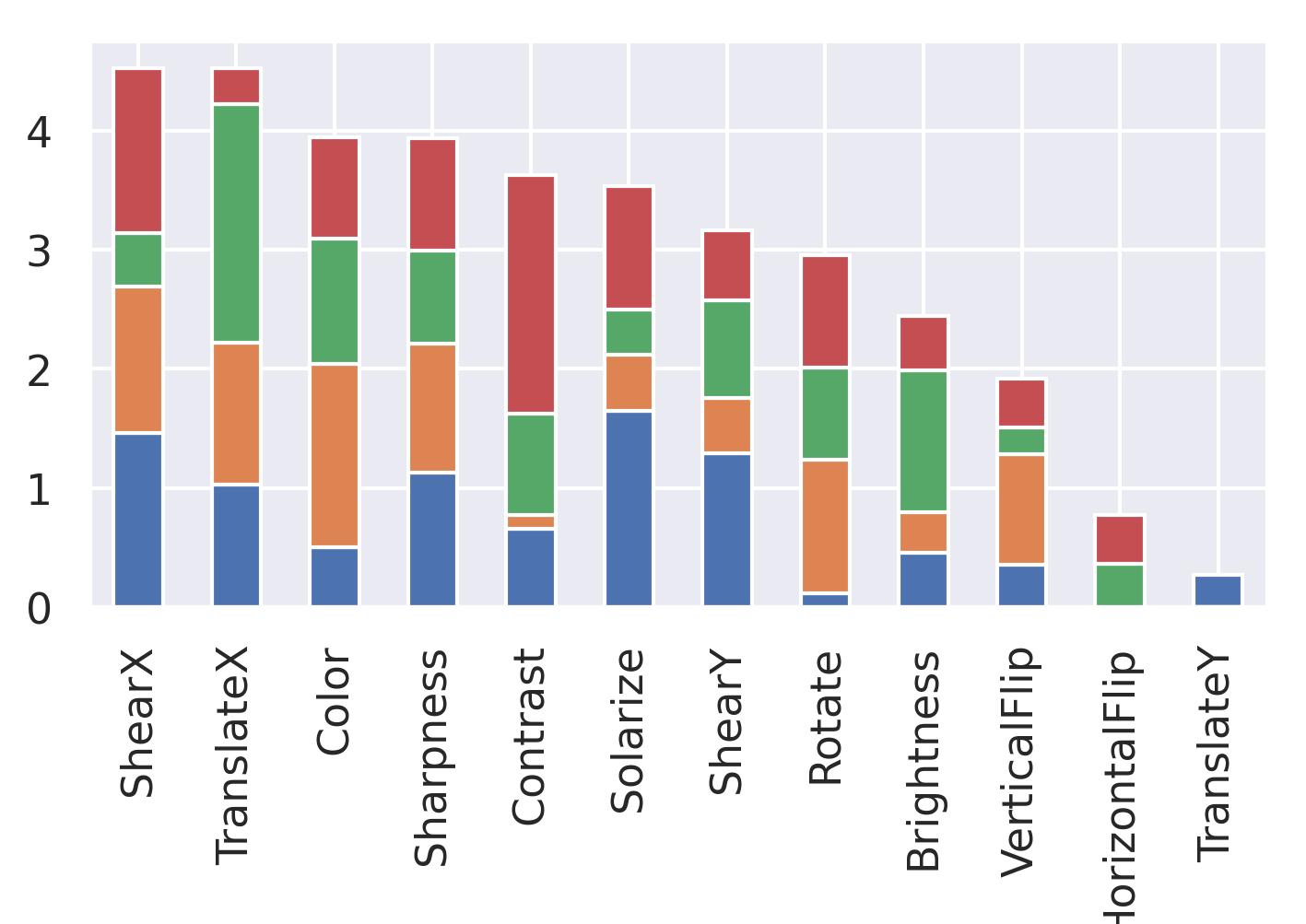} \\
  (a) Sensitivity SVHN  & (b) Importance SVHN \\
  \includegraphics[width=80mm]{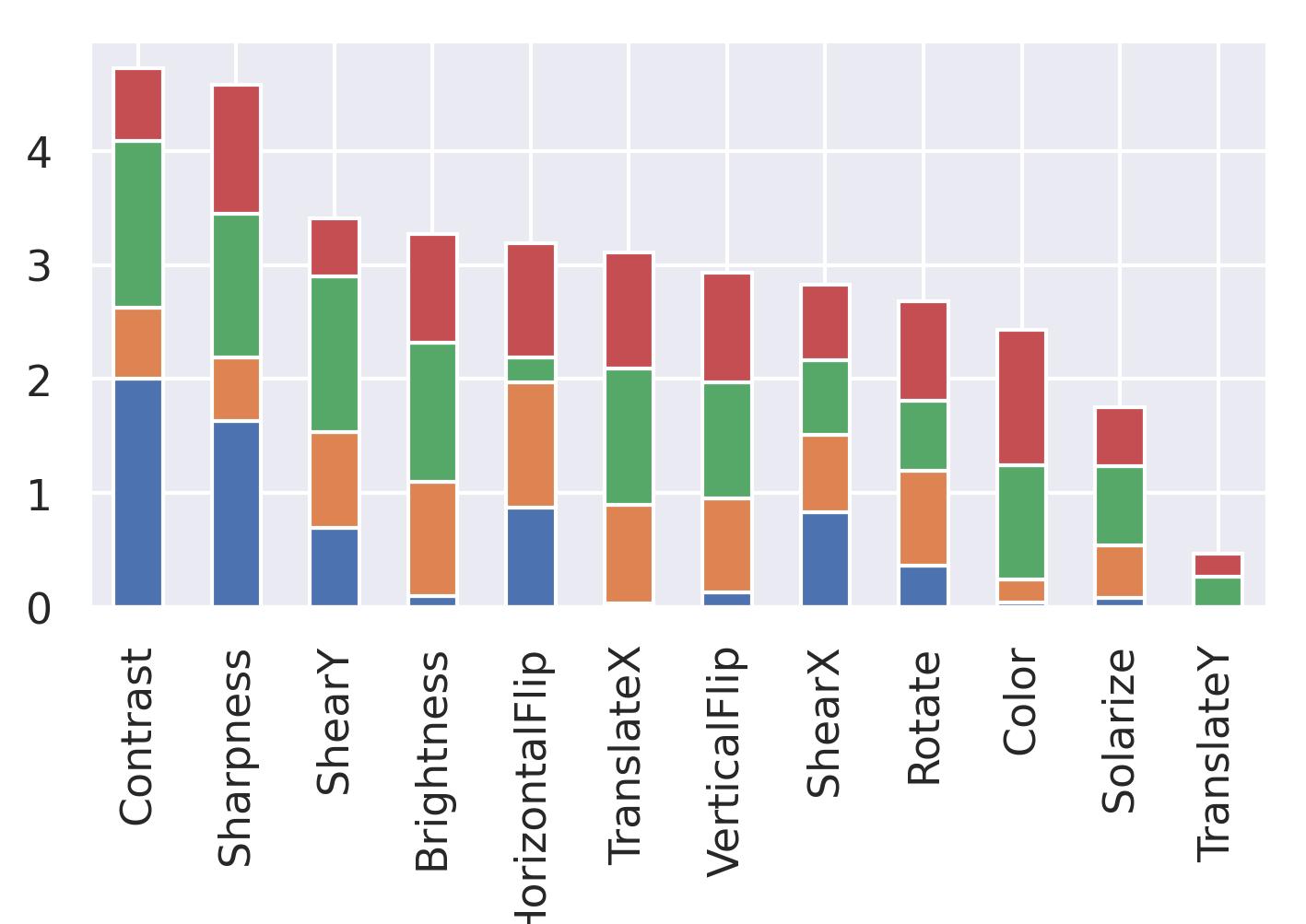} & \includegraphics[width=80mm]{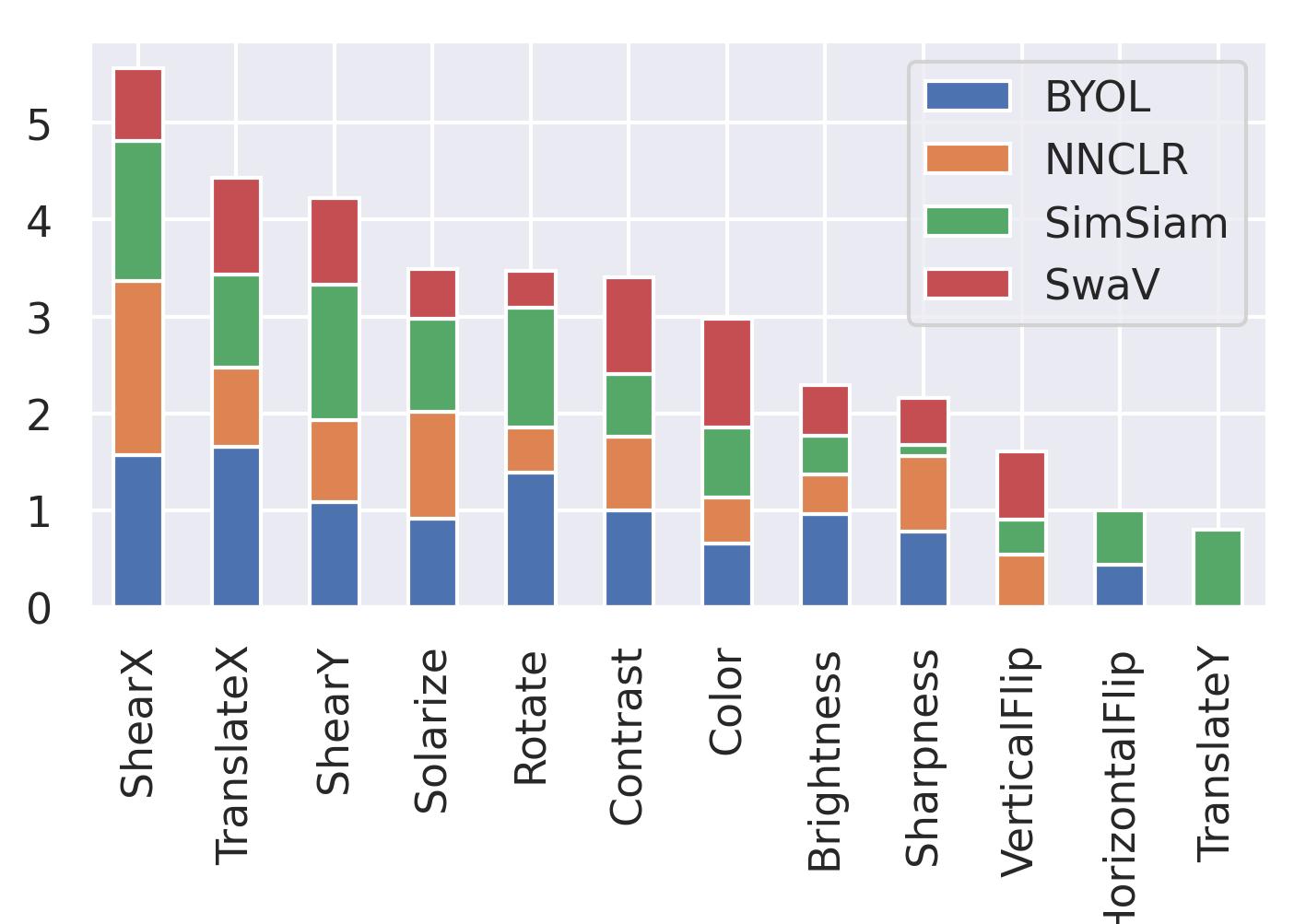} \\
  (c) Sensitivity Cifar10  & (d) Importance Cifar10 \\ [6pt]
\end{tabular}
\caption{Global view of augmentation sensitivity and importance}
\label{fig:op_sens_imp_all_stacked}
\end{figure}

Figure \ref{fig:op_sens_imp_all_stacked} demonstrates an overall view of augmentation importance and sensitivity for both batch sizes. From this Figure we can interestingly observe that both analyses follow nearly a similar trend and disclose similar information about each operator. Note that importance values show the more frequently used operators in the best policies, whereas, the sensitivity analysis accounts for the operators that make a bigger change in accuracy. Crucially, certain augmentation operators are found to be globally prevalent whereas others do not. In addition, the results of the four algorithms BYOL, NNCLR, SwAV and SimSiam, resembles a non-uniform distribution of sensitivity and importance values for the various operators. This supports the idea that specific augmentations lead to a stronger pretext task in the SSL algorithms. We would expect a more uniform distribution if the augmentations had no effect on the outcome. In our results we observe that in all cases, one or several augmentations dominate while others are very uncommon or are not even present in bar charts. 
Plots a and c of Figure \ref{fig:op_sens_imp_all_stacked} which are corresponding to the sensitivity analysis, illustrate that the algorithms are most sensitive to \textit{contrast} and \textit{sharpness} and least sensitive to \textit{translateY}. Both \textit{flip} operations have the second and third lowest overall sensitivity value, meaning that none of the algorithms appear to be overly sensitive to the \textit{flip} operation. \textit{TranslateX}, \textit{color}, \textit{brightness}, \textit{rotate}, \textit{solarize} and both \textit{shear} operators all have a medium sensitivity value. These findings suggest that all four SSL algorithms are more sensitive to the augmentations that are non-geometric transforms such as \textit{contrast} and \textit{sharpness}, and are less sensitive to the geometric transforms, such as \textit{translate} and \textit{flip}. 
From the stacked bar charts, it is also observable that certain operators are more influential in each dataset. It can be observed that SimSiam appears to be consistently sensitive to the \textit{brightness} operator in SVHN and to \textit{contrast} and \textit{shearY} in CIFAR-10, SwAV has a relatively high sensitivity to \textit{color} and \textit{rotate} in SVHN and several operators such as \textit{color} and \textit{sharpness} in CIFAR-10. BYOL consistently has a varied range of sensitivities to the different augmentation operators in SVHN, and in CIFAR-10, and the algorithm is most sensitive to \textit{contrast} and \textit{sharpness} operators. For NNCLR, we see that the experiments using SVHN are most sensitive to \textit{sharpness}. For CIFAR-10, there is not as clear of a pattern, however, \textit{horizontalFlip} is relatively showing a high sensitivity.

The augmentation importance bar charts in plots b and d of Figure \ref{fig:op_sens_imp_all_stacked} suggest that for both datasets \textit{shearX} and \textit{translateX} have appeared dominantly in top chromosomes, while \textit{verticalFlip}, \textit{horizontalFlip} and \textit{translateY} are the least applied operators. For SVHN, the remaining augmentations are more evenly distributed in terms of importance than Cifar10. We see overall the same augmentations for both datasets in between the lowest and high-importance augmentations, yet the ordering is slightly different. It is observed that \textit{sharpness} has a higher overall importance in SVHN than Cifar10, as does \textit{color}. \textit{Contrast}, \textit{solarize}, \textit{rotate}, \textit{shearY} and \textit{brightness} are all relatively similar in terms of overall importance for both datasets.  When considering the importance metrics for all four algorithms in Figure \ref{fig:op_sens_imp_all_stacked}, it is clear that specific augmentations are most important. For BYOL it is found that \textit{solarize} is the most important for SVHN and \textit{translateX} for Cifar10. Additionally, it can be seen that both \textit{shear} operations are consistently of relatively high importance for both datasets with BYOL. For NNCLR, \textit{color} is most important with SVHN and \textit{shearX} is most important with Cifar10. The most important operators for SimSiam are \textit{translateX} for SVHN and \textit{shearX} and \textit{shearY} for Cifar10.  Lastly, for SwAV \textit{contrast} is the most important augmentation for both datasets.

\subsection{Landscape Analysis}
\begin{figure}
\centering
 \begin{tabular}{cccc}
 \includegraphics[width=41mm]{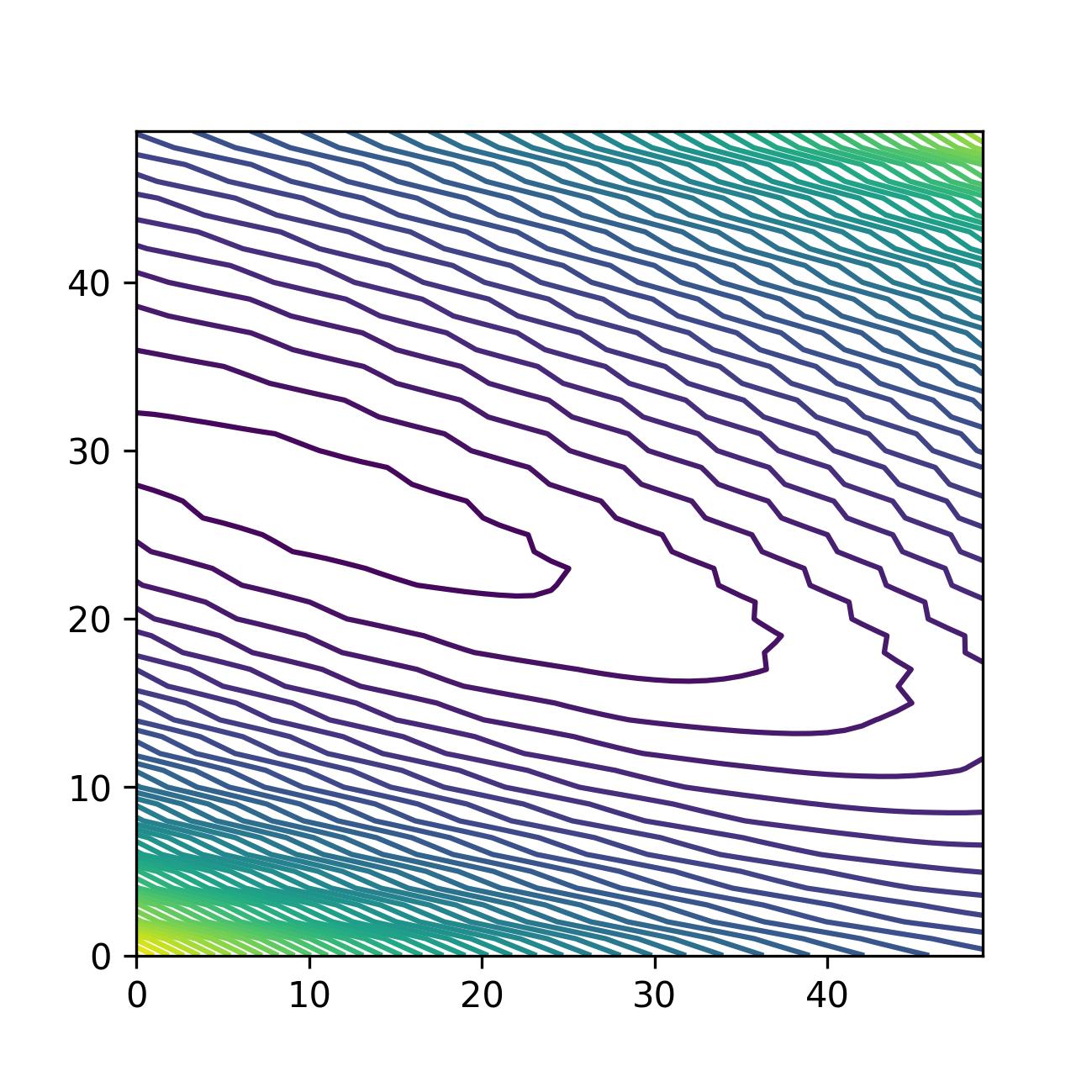} & \includegraphics[width=41mm]{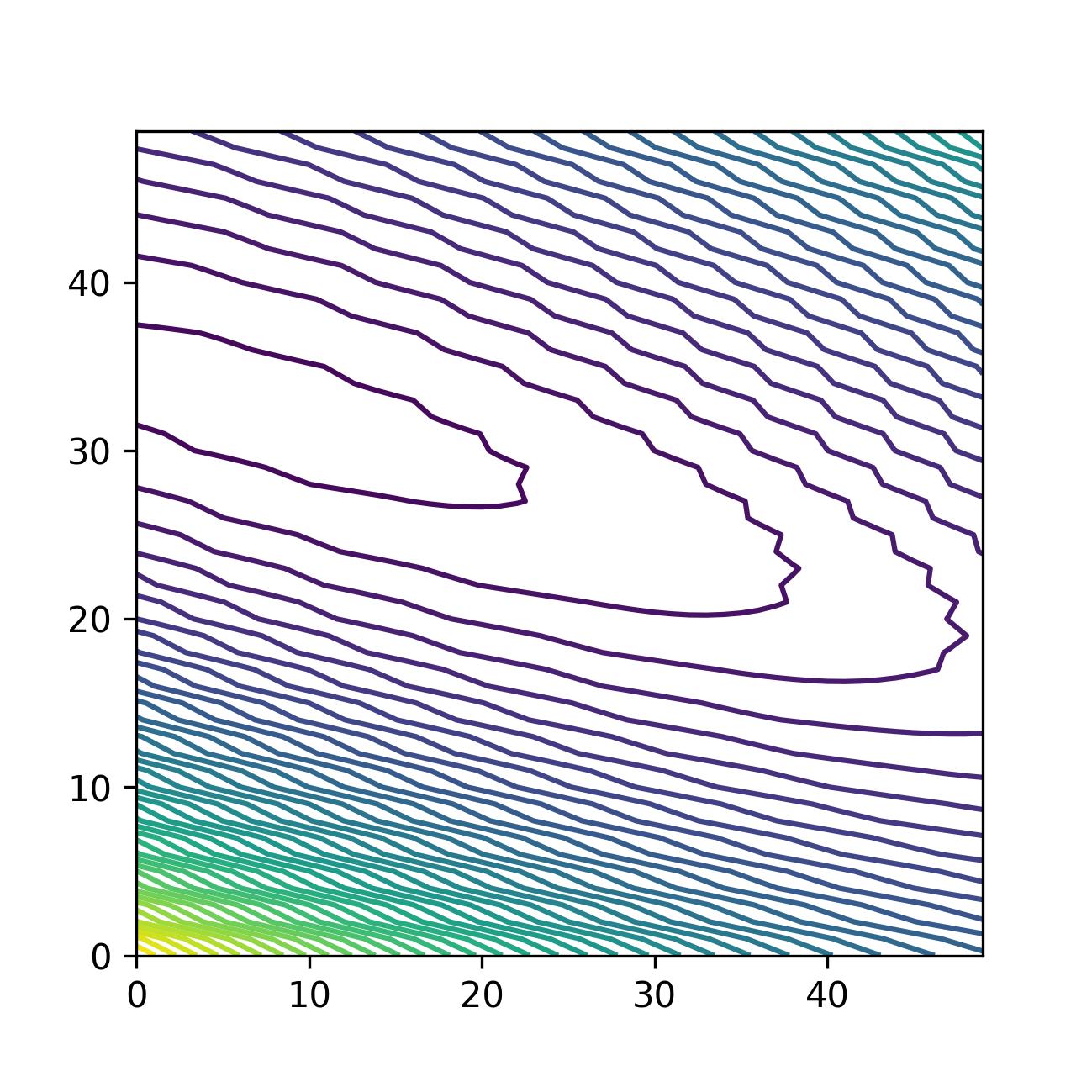} &
  \includegraphics[width=41mm]{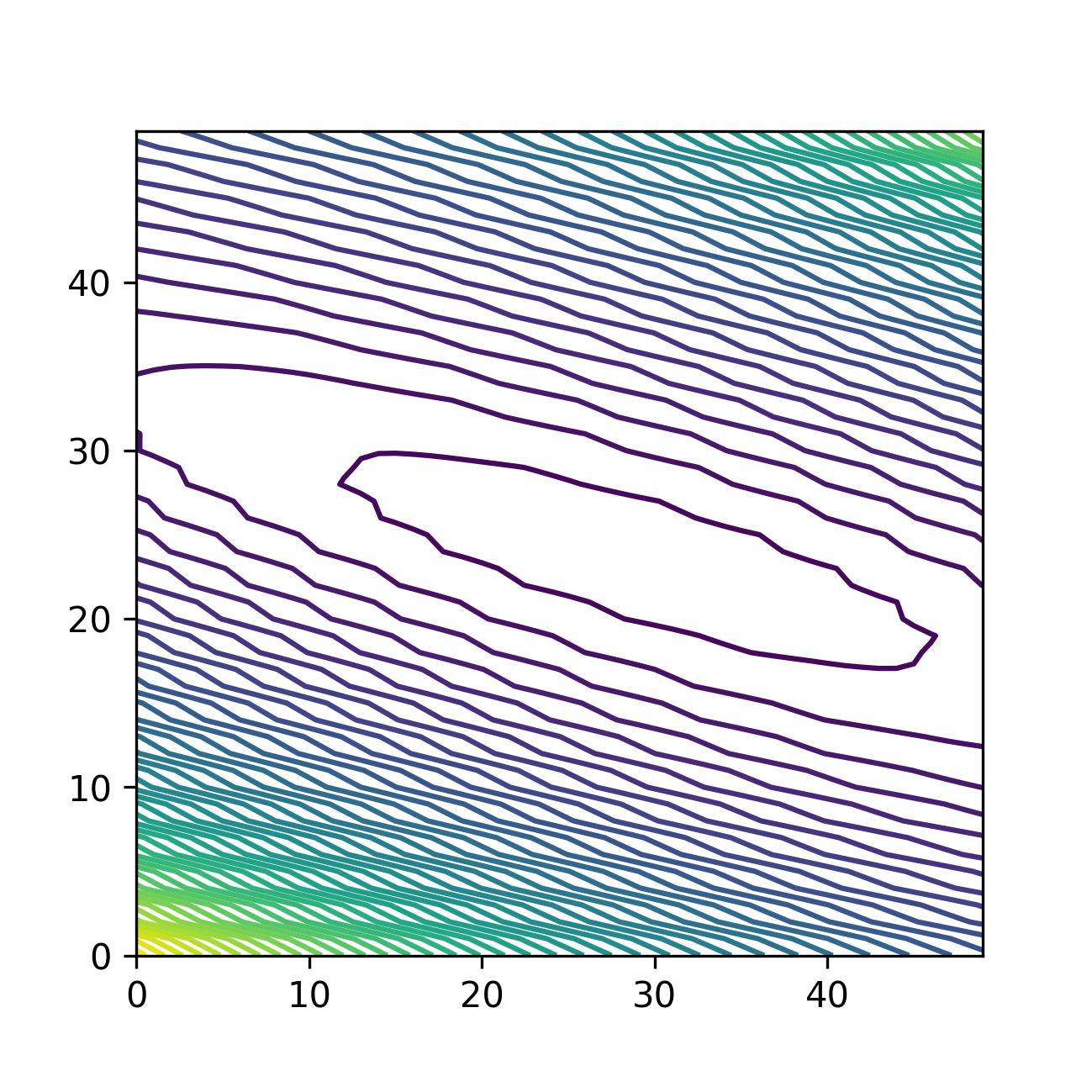} & \includegraphics[width=41mm]{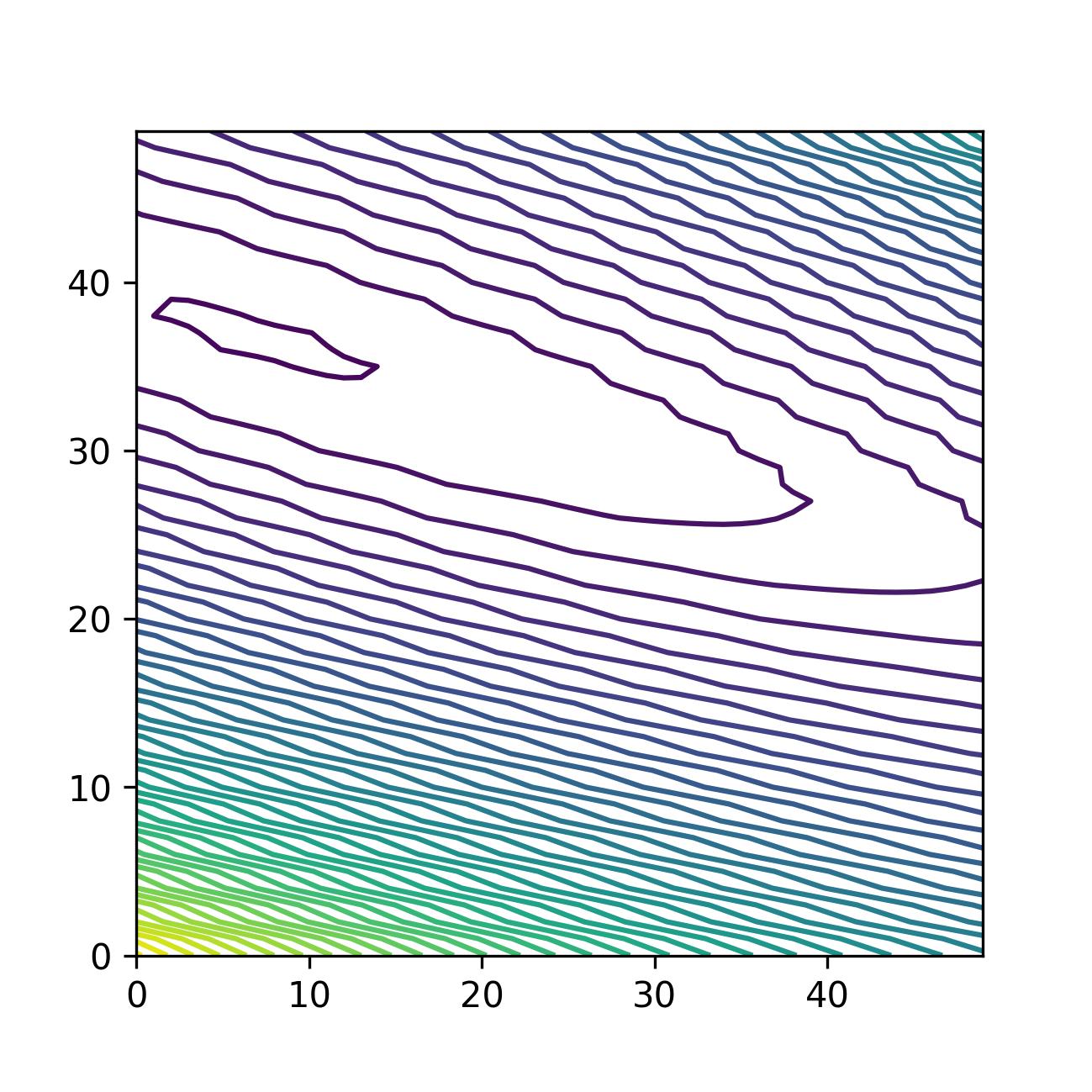} \\
  \includegraphics[width=41mm]{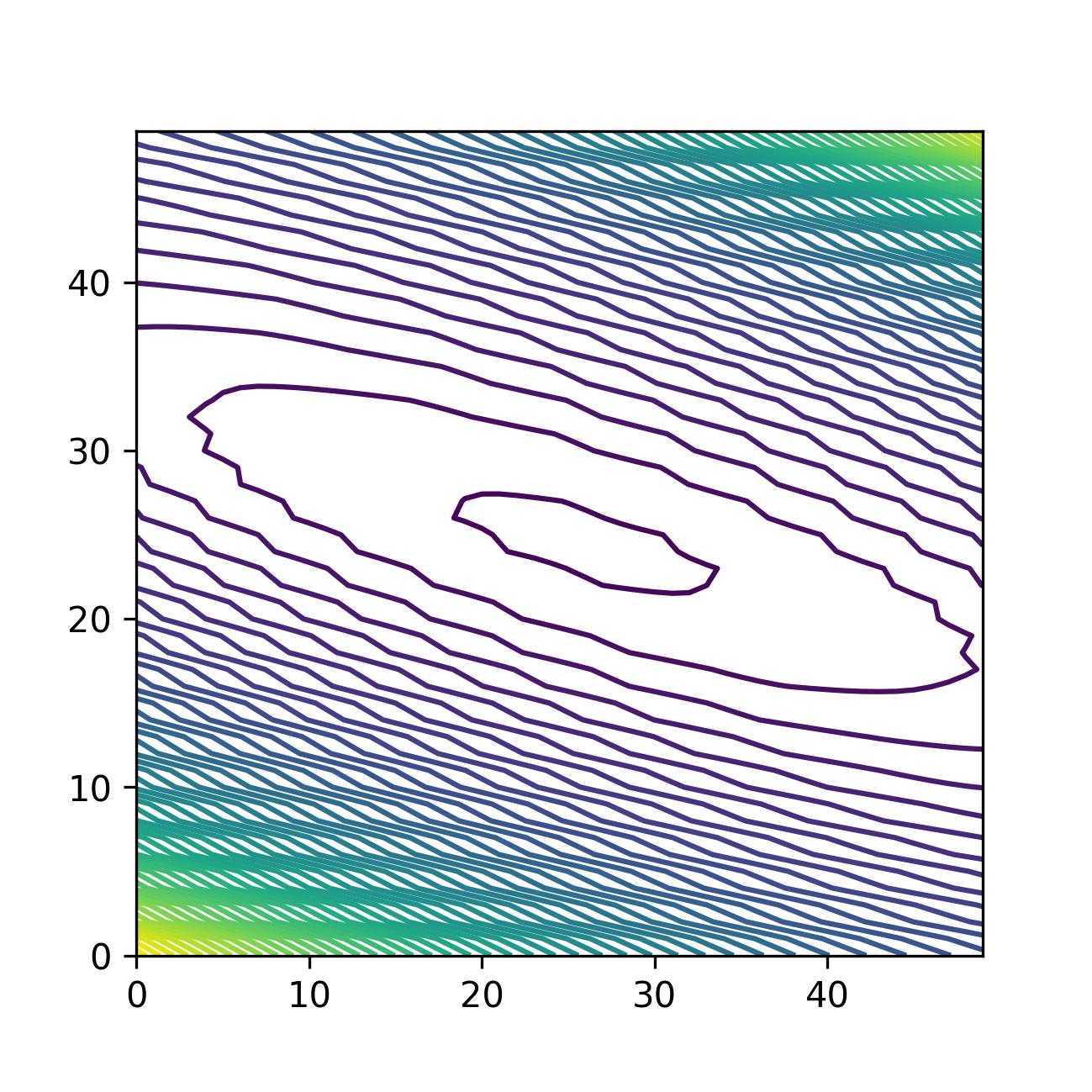} & \includegraphics[width=41mm]{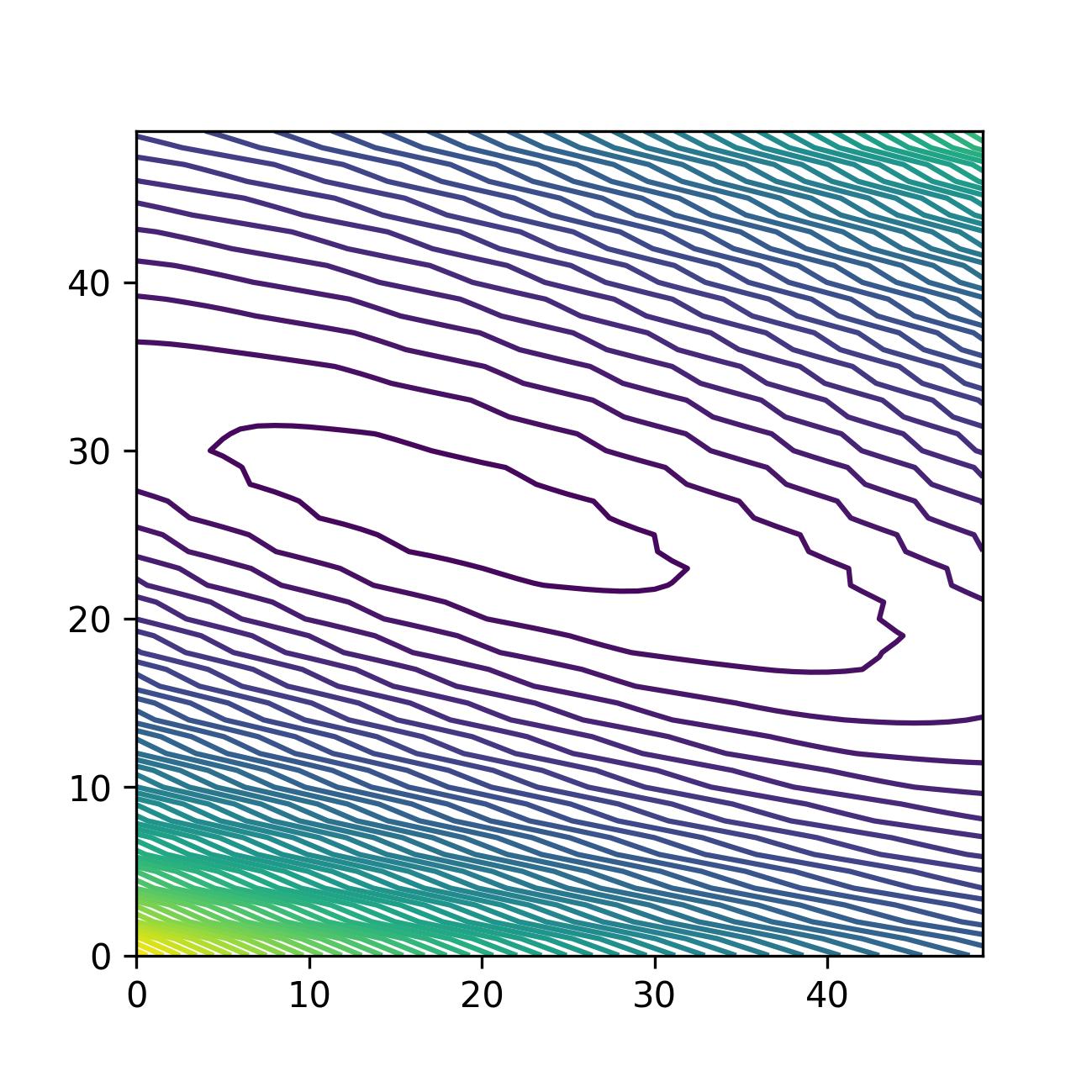} &
  \includegraphics[width=41mm]{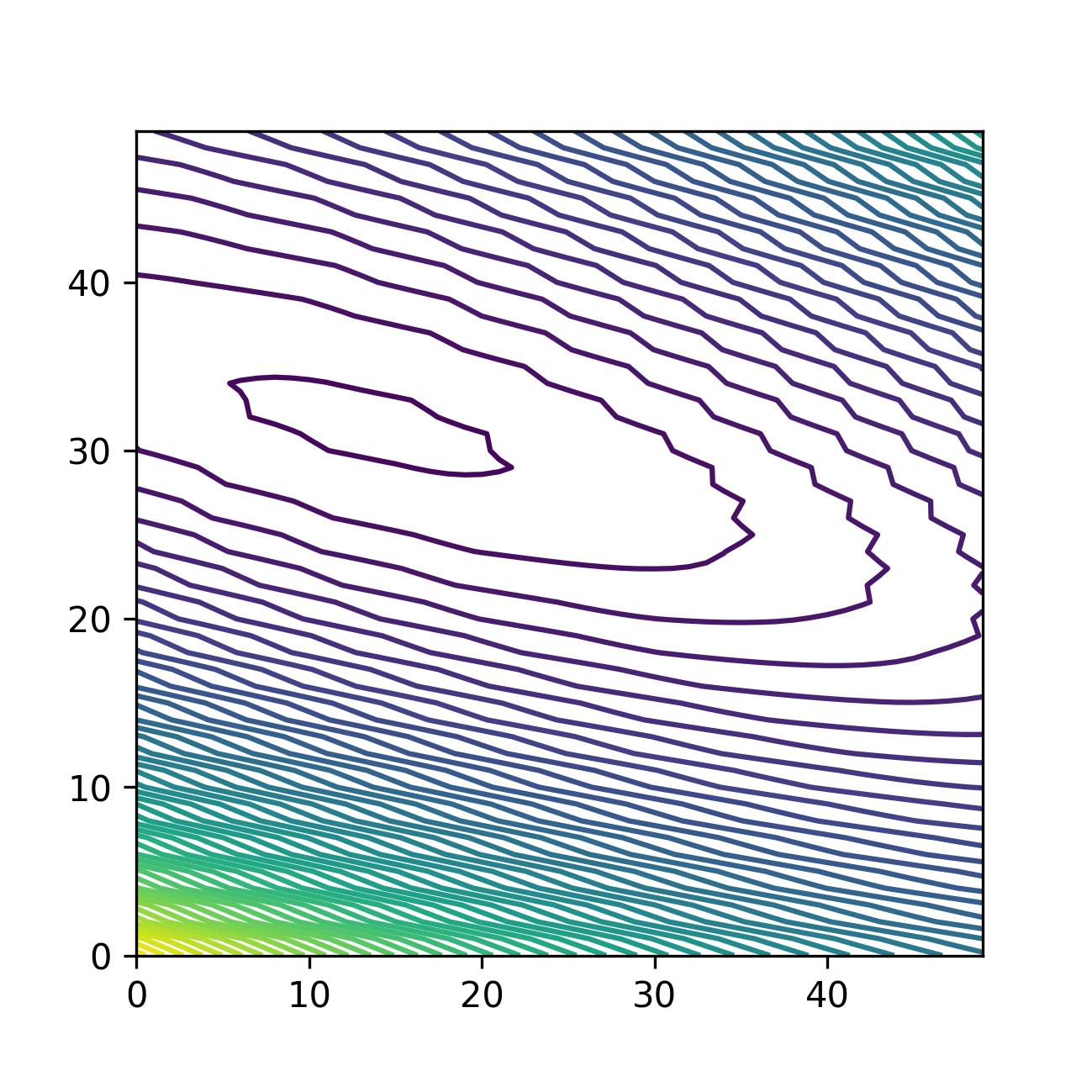} & \includegraphics[width=41mm]{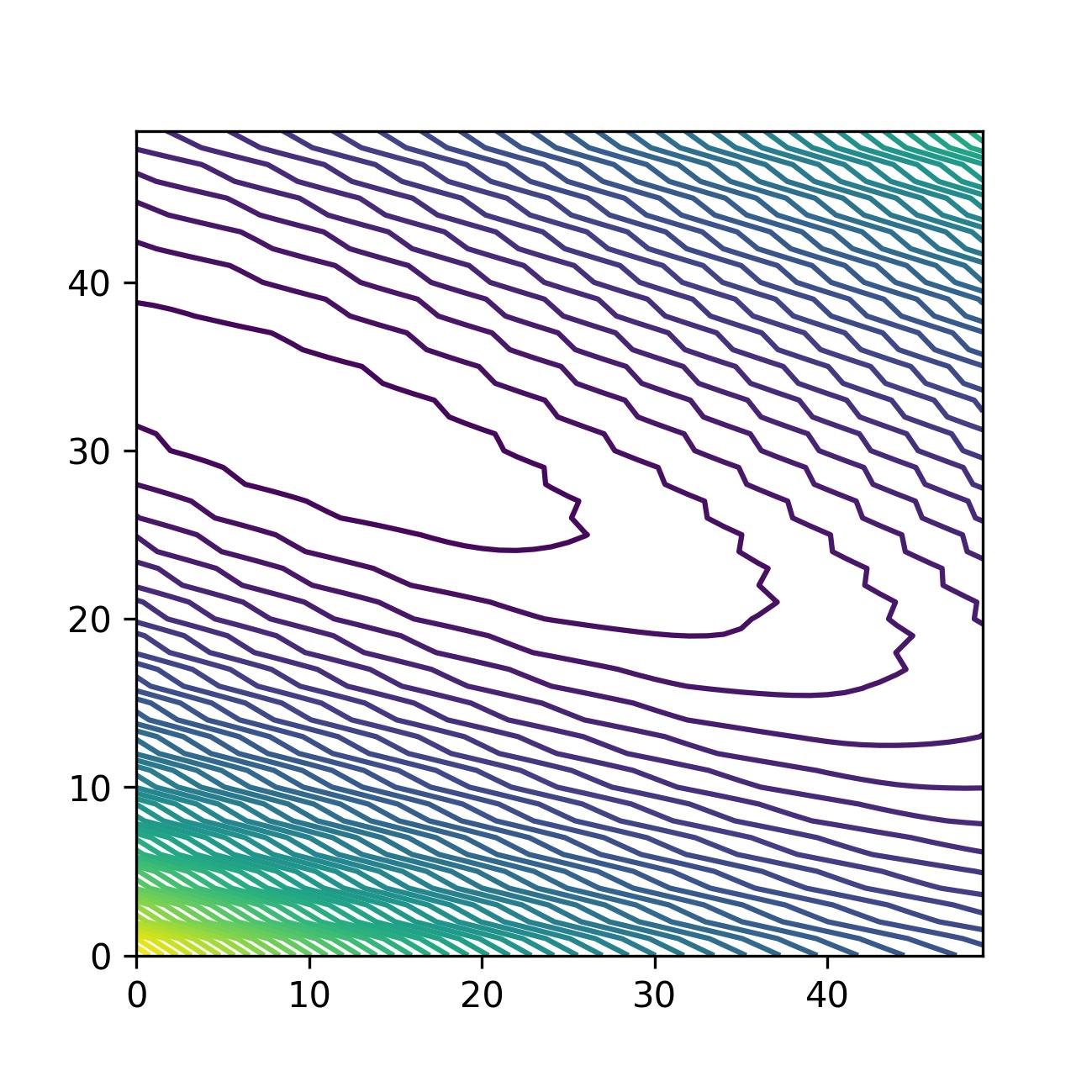} \\
  (a) BYOL & (b) NNCLR & (c) SimSiam & (d)  SwAV \\
\end{tabular}
\caption{Comparing the loss landscape of the downstream models, using SSL methods with a batch size of 32 (top row) and batch size of 256 (bottom row). We see that the smaller batch size in the SSL pretraining leads to a less sharp minimization in the downstream task.}
\label{fig:loss_landscape}
\end{figure}

To better compare the SSL algorithms and in order to measure the effect of batch size in our experiments we utilize a strategy for analyzing the loss of the resultant downstream models, known as loss landscape analysis. Loss Landscape analysis aims to provide a deeper understanding of the optimized model parameters and its context in its respective parameter space by slightly perturbing the model parameters around the optimized parameterization. This provides a \textit{landscape} of the loss function for the given parameter space, allowing  for insights into how convex or chaotic the loss space around the solution is. Producing a loss landscape for a given model architecture and problem is not a trivial task and many different approaches exist for doing so. This work opts to employ a filter normalization approach presented by Li \textit{et al.} \cite{li_visualizing_2018} to visualize the loss landscape. 
The filter normalization loss landscape visualization was chosen for this work as it was found to accurately capture the local sharpness and flatness of minimizers. Understanding the local geometry of the loss landscape is an essential component for understanding the behaviour of the loss function in the problems at hand. As discussed in \cite{li_visualizing_2018}, a key component of the filter wise normalization technique is a random vector technique used by \cite{Goodfellow2014NEC}, \cite{Jiwoong2016LossSurfaces}. In this technique two random vectors $\delta$ and $\eta$ are sampled from a random Gaussian distribution, as well as a center point $\theta*$, where $\theta*$ is the optimized parameter configuration. Then a 2d contour is generated by plotting the function\ref{eq:random_dirction}:
    \begin{equation}
    \label{eq:random_dirction}
        \begin{aligned}
           f(\alpha, \beta)  =  L(\theta* + \alpha\delta + \beta\eta)
        \end{aligned}
    \end{equation}
where $\alpha$, and $\beta$ are the respective lengths of random vectors $\delta$ and $\eta$. 
The key component of this approach which differs from earlier random vector approaches is that, the sampled directions are filter-wise normalized. That is the random vectors are scaled to the filters within the CNN to ensure that the area covered by the random vectors are not too small or too large for the set of parameters.
Mainly, it was found that the results with Cifar-10 were more interesting in terms of effect of batch size. For this reason we choose to further explore the loss landscapes of the experiments on Cifar-10 dataset. For all four SSL algorithms, NNCLR, BYOL, SimSiam and SwAV and the two batch sizes 32, and 256, we use the filter normlaization technique with a $\alpha$, and $\beta$ both set to $10$ and a sample size of $50$ resulting in a $50x50$ loss landscape.  Li \textit{et al.} \cite{li_visualizing_2018} found that with the filter normalization method it was possible to visualize how batch size impacts minima sharpness, large batches were found to produce visually sharper minima. We visualize the loss landscape of the best solutions found by our evolutionary search mechanism. Remarkably, in all four SSL algorithms we notice a difference in sharpness when training with a batch sizes of 32 and 256 in the pretext task and keeping a fixed batch size of 32 in the downstream task. As illustrated in Figure \ref{fig:loss_landscape}, we observe that BYOL, NNCLR and SimSiam all have converged to sharp minima in the downstream task when using a batch size of 256 in the pretext task, and a visibly flatter minima when using a batch size of 32 in the pretext task, the opposite result is observed in SwAV.

\section{Conclusion}
Our contribution in this paper is two-fold; On one hand, we focused on optimizing the performance of SSL algorithms by finding the best augmentation operators. To this end, we proposed an approach based on evolutionary optimization to automatically find the optimal augmentation operators and their intensities in order to maximize the accuracy of downstream task. In this regard, we evaluated and compared the performance of four SOTA SSL algorithms optimized by our proposed method. Our results indicated that our proposed method boosts the accuracy of downstream classification task. On the other hand, we proposed algorithms for explaining the optimized solutions in order to analyze and find the impact of augmentation operators applied in the pretext task of SSL algorithms. Accordingly, we designed explainability experiments to elicit the most influential augmentation operators using two standard visual datasets. We further analyzed the impact of different parameters including batch size and epoch number, and visualized loss landscapes to better understand the best augmentation policies and best performing models. To the best of our knowledge this work is the first attempt to optimize hyper-parameters used in the pretext task for self-supervised learning algorithms. Our results confirm that SSL algorithms are sensitive to the choice of augmentation policies. We also demonstrated that our proposed evolutionary augmentation optimization method can find the solutions that lead to performance enhancement of SSL algorithms by efficiently searching the augmentation policy space. 

\section*{Conflict of Interest Statement}
The authors declare that the research was conducted in the absence of any commercial or financial relationships that could be construed as a potential conflict of interest.

\section*{Funding}
This research has been supported by the Natural Sciences and Engineering Research Council of Canada.

\section*{Data Availability Statement}
The datasets analyzed in this study can be found at [https://www.cs.toronto.edu/~kriz/cifar.html] and [http://ufldl.stanford.edu/housenumbers/].

\bibliographystyle{Harvard} 
\bibliography{References}
\end{document}